\RequirePackage{amsmath}
\documentclass[runningheads]{llncs}
\usepackage{graphicx}

\usepackage{tikz}
\usepackage{comment}
\usepackage{amsmath,amssymb} %
\usepackage{color}

\usepackage{booktabs}
\usepackage{textcomp} %
\usepackage{setspace} %
\usepackage{microtype}
\usepackage{hyperref}
\usepackage{attrib}
\usepackage{epigraph}
\usepackage{enumitem}

\usepackage[rightcaption]{sidecap}

\newcommand{\etal}{\textit{et al.\ }}

\usepackage[width=122mm,left=12mm,paperwidth=146mm,height=193mm,top=12mm,paperheight=217mm]{geometry}

\DeclareRobustCommand\onedot{\futurelet\@let@token\@onedot}
\def\@onedot{\ifx\@let@token.\else.\null\fi\xspace}

\begin{document}
\pagestyle{headings}
\mainmatter
\def\ECCVSubNumber{2473}  %

\title{Learning to Factorize and Relight a City}

\titlerunning{Learning to Factorize and Relight a City}
\author{Andrew Liu\inst{1}
\and
Shiry Ginosar\inst{2}
\and
Tinghui Zhou\inst{3} \and \\
 Alexei A. Efros\inst{2} \and Noah Snavely\inst{1}} 
\authorrunning{A. Liu et al.}
\institute{
Google\inst{1} \; \; UC Berkeley\inst{2} \; \; Humen, Inc.\inst{3} 
}
\maketitle

\vspace{-0.1in}
\begin{abstract}
We propose a learning-based framework for disentangling outdoor scenes into temporally-varying illumination and permanent scene factors. Inspired by the classic intrinsic image decomposition, our learning signal builds upon two insights: 1) combining the disentangled factors should reconstruct the original image, and 2) the permanent factors should stay constant across multiple temporal samples of the same scene. To facilitate training, we assemble a city-scale dataset of outdoor timelapse imagery from Google Street View, where the same locations are captured repeatedly through time. This data represents an unprecedented scale of spatio-temporal outdoor imagery. We show that our learned disentangled factors can be used to manipulate novel images in realistic ways, such as changing lighting effects and scene geometry. Please visit \url{factorize-a-city.github.io} for animated results.

\end{abstract}
\vspace{-0.25in}
\begin{figure}[h]
\centering
\includegraphics[width=\linewidth]{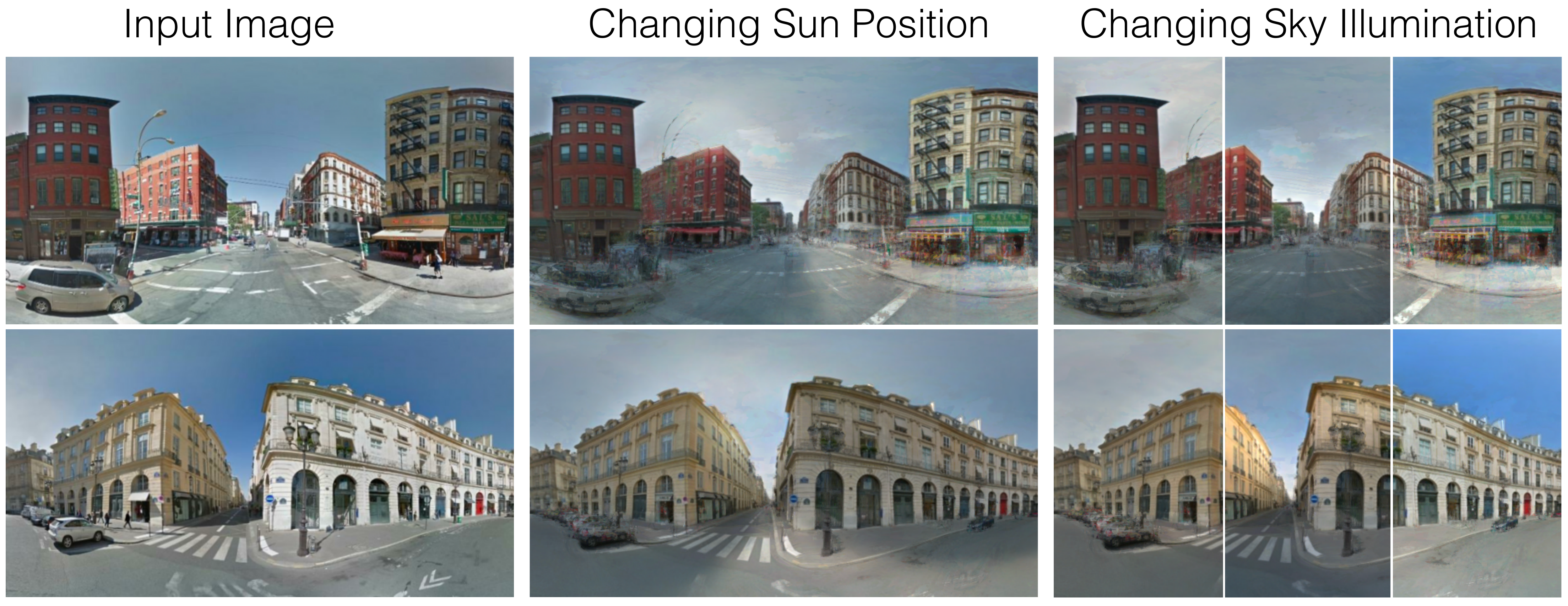}
\vspace{-.3in}
\caption{We learn to disentangle temporally-varying scene factors from permanent ones.
We can manipulate the learned factors
to relight scenes, e.g., by editing sun position and sky conditions. While we train our model on panoramas of NYC~\textit{(top)}, it generalizes at test time to images of other cities such as Paris~\textit{(bottom)}.
}
\label{fig:teaser}
\end{figure}
\vspace{-0.25in}

\section{Introduction}

\vspace{-.05in}
\epigraph{\textit{``The city of Sophronia is made up of two half-cities... One of the half-cities is permanent, the other is temporary."}}{\sc{--- Italo Calvino,} \em{Invisible Cities}}
\vspace{-.05in}

Imagine taking an image from every possible location on Earth at every possible time instant throughout history.  Adelson and Bergen called this hypothetical construct the {\em plenoptic function}~\cite{AdelsonBergen1991}.  In practice, of course, it would be impossible to capture or store such a massive dataset. 
Yet, the data must also be highly redundant and compressible.   
There will be many images of the same view with slightly different illumination, many images capturing different places under the same conditions, etc. In other words, each image within this hypothetical dataset should have a low intrinsic dimensionality. Rather than store all pixels, we could instead store a small number of intrinsic, disentangled factors representing scene geometry, illumination conditions, etc.---if only we knew what those parameters were and how to reconstruct an image from them.

In this paper, we ask whether we can learn such a lower-dimensional representation from a sparse sampling of the plenoptic function on the scale of an entire city. Until recently, 
large-scale visual data that varies both in space and, separately, in time was difficult to obtain.  Fortunately, there have been systematic efforts to capture the world through projects like Google Street View (GSV). While GSV is known for its worldwide coverage, it has also accumulated many samples of the world over time, powering features like Street View Time Machine (GSV-TM). However, GSV-TM still represents an extremely sparse sampling of the plenoptic function.

We use GSV to learn to factor a city's worth of outdoor panoramas into a single low-dimensional representation.
In particular, we organize a large set of historical GSV panoramas of New York City into \emph{assembled timelapses} at 100,000 fixed locations captured over time. These enable us to train an unsupervised model to disentangle two latent factors: illumination factors that vary over time, and geometric scene properties that are more permanent.

Once we learn a disentangled set of latent factors, we can synthesize missing data in our incomplete sampling of the plenoptic function 
by simply swapping or modifying the underlying factors. 
As illustrated in Fig.~\ref{fig:teaser}, our learned factorization can generate synthetic images of the same scene with completely novel illumination. Our disentangled factors are flexible enough to relight test scenes from a single panorama and can even be applied to entirely new cities like Paris.

\section{Related Work}
\noindent \textbf{Intrinsic Images.}
Decomposing images into their underlying components 
is a well-studied problem~\cite{barrow1978}. For instance, the classic intrinsic images problem describes images as a combination of \textit{reflectance} (i.e., scene albedo), and \textit{shading} (effects induced by lighting)~\cite{Adelson_reflectance}. This problem is underconstrained as there are an infinite number of possible solutions for a single image. However, the regularities in natural scenes and lighting conditions allow for priors on the decomposition.
While such priors can be manually crafted~\cite{BarronTPAMI2015}, many recent methods attempt to learn priors from data, using full supervision from synthetic data~\cite{li2018cgintrinsics}, sparse supervision from human annotations~\cite{Bell:2014:IIW:2601097.2601206,zhou2015intrinsic}, or self-supervision from synthetic models~\cite{janner2015}.
Yet another kind of supervision comes from \emph{timelapse videos}~\cite{BigTimeLi18}, which feature image sequences with constant reflectance but varying illumination. Such work hearkens back to classic work on deriving intrinsic images from image stacks~\cite{weiss2001intrinsics}, and is an inspiration for our work. However, while intrinsic image methods allow for editing reflectance or shading for a specific image, they use high-dimensional \emph{pixel-level} descriptions of lighting that are not transferable across scenes. 
In our case, our model learns an illumination descriptor that can be meaningfully transferred from one image to another, e.g., to relight an image with an illumination from a completely different scene. Such ``mix-and-match'' capabilities are beyond the power of standard intrinsic images.

\medskip
\noindent
\textbf{Inverse Graphics.} 
An alternative way to factor visual appearance 
is via 3D reconstruction of the scene 
into underlying physical components like 3D shape, materials, and lighting. Such methods have been successful in several specific domains, including faces~\cite{sengupta2018sfsnet}, single objects~\cite{birdsbirdsbirds,VON}, or indoor scenes trained from synthetic data~\cite{sengupta2019neural,li2020inverse}. 
3D reconstruction has also been used explicitly as a preprocess to aid in modeling visual appearance~\cite{TimelapseMiningSIGGRAPH15,Laffont_2015_ICCV,PGZED19,Meshry_2019_CVPR}. Most relevant to us are Martin-Brualla~\textit{et al.}~\cite{TimelapseMiningSIGGRAPH15}, who organized millions of internet photos into a dense 3D and temporal reconstructions, and Meshry~\textit{et al.}~\cite{Meshry_2019_CVPR}, who employed a dense 3D reconstruction with a neural rendering pipeline to synthesize scene appearances. However, explicit 3D reconstruction methods require hundreds of images to create a 3D model and cannot generalize to novel test-time scenes.
In contrast, we choose to handle geometry implicitly---allowing us to holistically learn to disentangle factors across many scenes composed of a few images each, and then generalize to novel settings, even single images. 

Some recent inverse graphics methods learn to infer shape, appearance, and materials for new outdoor scenes, not just scenes observed during training. Yu and Smith train on multi-view stereo data using a physics-based inverse graphics model, and can infer explicit scene properties for novel test images, enabling relighting tasks~\cite{yu19inverserendernet}. Our work achieves a similar capability, but relies on a more implicit representation of geometry and illumination that can be learned solely from timelapse data, without requiring depth or surface normals during training.

\medskip
\noindent \textbf{Timelapse and Webcam Data.}
Timelapses are a popular source of data for capturing 
time-related effects. 
Applications include intrinsic images~\cite{weiss2001intrinsics,BigTimeLi18}, 
scene-specific factorizations via physical shading models~\cite{ftlv}, 
illuminant transfer~\cite{lalondewebcam}, 
analysis of worldwide temporal variations~\cite{amos_webcam},
motion denoising~\cite{Rubinstein11Motion}, 
learning temporal object transformations~\cite{bergtimelapse}, 
and weather attribute manipulation~\cite{Laffont14}.
However, prior work is limited by the variety and size of available data. The largest existing set of standard webcam data is the 
AMOS dataset of Jacobs \textit{et al.}~\cite{amos_webcam}, 
which archived 29,445 webcams and %
95 million images.
BigTime~\cite{BigTimeLi18} uses a
much smaller set of 6,500 images from 195 timelapse sequences. 
Both datasets sample \textit{time} much more densely than \textit{space}.
In contrast, we leverage the vast amounts of data from Google Street View to create \textit{assembled timelapses} of the same location captured at different times, across a large number of locations. This allows us to collect an order of magnitude more data than previously published~\cite{amos_webcam}. We additionally note that data collection from Street View scales more easily than~\cite{amos_webcam} which requires crawling the internet for webcam streams.

\medskip
\noindent \textbf{Learning from Street View.}
Google Street View (GSV), a large dataset of images sampling much of the world's streets, represents a compelling source of data for computer vision research.
Researchers have utilized Google Street View images to learn about visual elements~\cite{parisparis} or historical architectural styles~\cite{linking2015iccp} specific to certain cities like Paris, 
to predict non-visual city attributes~\cite{streetscore,cityforensics,socioeconomiccars},
for localization~\cite{Gronat_2013_CVPR}, 
or to understand the relationship between satellite imagery and street-level views~\cite{haysoverhead}.
In our work we use historical GSV 
Time Machine 
imagery to observe how the world changes over time by assembling timelapses for a large number of locations. 
Such a large, comprehensive dataset is key to our unsupervised approach for learning to factor illumination from scene geometry.

\section{Google Street View Time Machine Data}

\begin{figure*}[t!]
\centering
\includegraphics[width=\linewidth]{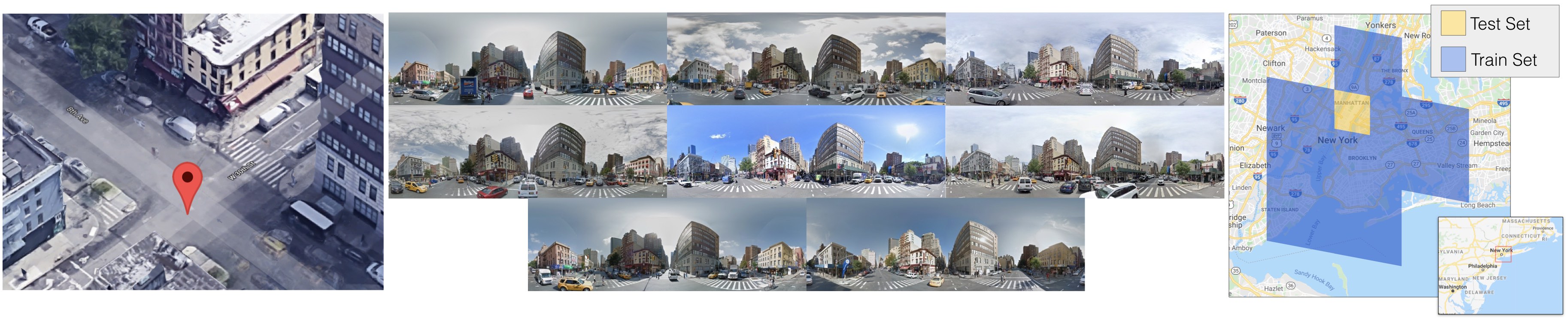}
\vspace{-.3in}
   \caption{\textit{Left:} A Manhattan intersection. 
   \textit{Center:} Multiple Google Street View panoramic captures of this intersection 
forms an \textit{assembled timelapse} stack. 
\textit{Right:} The train and test split over the greater NYC area. Training stacks are drawn from the blue region, and test stacks from the yellow region.
   }
\label{fig:map-stack}
\end{figure*}

\label{sec:data}
Google Street View (GSV) hosts an amazing quantity of panoramas capturing street scenes worldwide.
Because GSV repeatedly captures many places over time, it can be treated as a sparse, imperfectly aligned, and irregularly-sampled collection of timelapse videos. These historical images are saved as part of the GSV-Time Machine (GSV-TM), which we mine to collect our dataset.

We focus on 
New York City, due to the richness of NYC scenes and the relative wealth of 
data. To assemble timelapses, we  
collect panoramas within 
NYC along with their timestamps and camera poses 
in a geographic coordinate system~\cite{klingner}. 
We greedily cluster nearby panoramas into sets of eight, which we refer to as \emph{stacks}. The region we use and an example stack are shown in 
Fig.~\ref{fig:map-stack}. %

From the area shown in Fig.~\ref{fig:map-stack} (right) we collect $\sim$100K assembled timelapse stacks for training (comprised of 800K individual panoramas stitched from 10 million captures) and 16K test stacks.
We crop the sky and ground regions
such that our final panoramas are $960 \times 320$. These sRGB panoramas can optionally be gamma-corrected before further processing.

\section{Method}
Our goal is to discover a low-dimensional representation of the world where temporally varying effects, such as different illumination conditions, are disentangled from permanent objects, such as buildings and roads.

\begin{figure*}[t!]
    \centering
    \includegraphics[width=\linewidth]{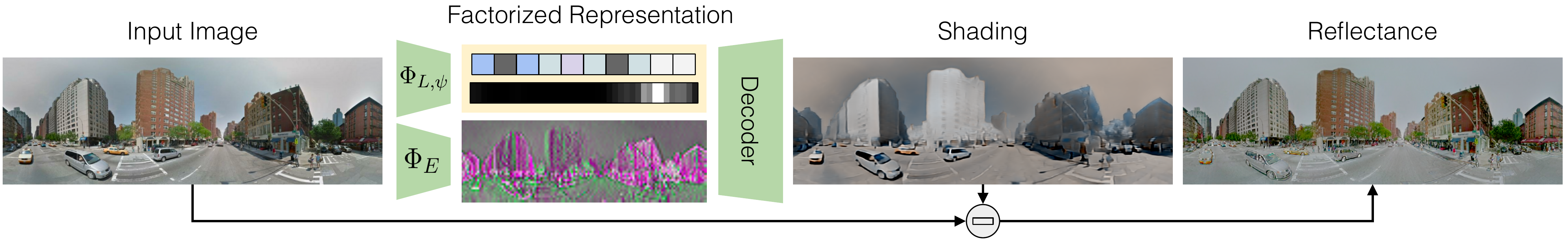}
    \vspace{-.3in}
    \caption{\textbf{Disentangling a single image.} At test time, we \textit{encode} a single image into disentangled time-varying and permanent factors. We train with the constraint that shading and reflectance images can be \textit{decoded}  from this learned factored representation.}
    \label{fig:main_test}
\end{figure*}

One form of disentanglement
is \emph{intrinsic images}, a per-pixel decomposition into reflectance and shading images. However, such a disentangled representation is very low-level---a particular shading image cannot be used to relight a different scene. Instead, we seek to encode an image into higher-level latent factors capturing scene and illumination properties described above, as illustrated in Fig.~\ref{fig:main_test}. How can we find such a factorization? Our insight is that we should still be able to \textit{decode} intrinsic images from our factored representation, as illustrated on the right side of Fig.~\ref{fig:main_test}. The decoded reflectance and shading images should recombine to form the original image, providing us with an autoencoder-style method for learning our high-level factorization~\cite{janner2015}. However, such an image reconstruction framework alone would provide a very weak supervision signal. Our second insight is to learn from huge numbers of \textit{timelapse stacks} mined from GSV-TM. Within such stacks, we assume the scene factors to be constant. This insight is inspired by the work of Li and Snavely, who learn intrinsic images from timelapse videos~\cite{BigTimeLi18}. In our case we learn a high-level factorization that enables more powerful capabilities.

\subsection{Encoder-Decoder Architecture}
\label{sec:encoder}
Fig.~\ref{fig:main_test} shows our encoder-decoder architecture with its learnt factored representation.
Given an image, our encoders produce latent factors, capturing various temporal and permanent effects, that can be decoded to a log-shading intrinsic image. We use the intrinsic images equation
($\log(\mathrm{Reflectance}) = \log(\mathrm{Image}) - \log(\mathrm{Shading})$)
to compute a reflectance image by subtracting the temporally varying effects, represented by the shading image, from the original image.
\begin{figure*}[t!]
    \centering
    \includegraphics[width=\linewidth]{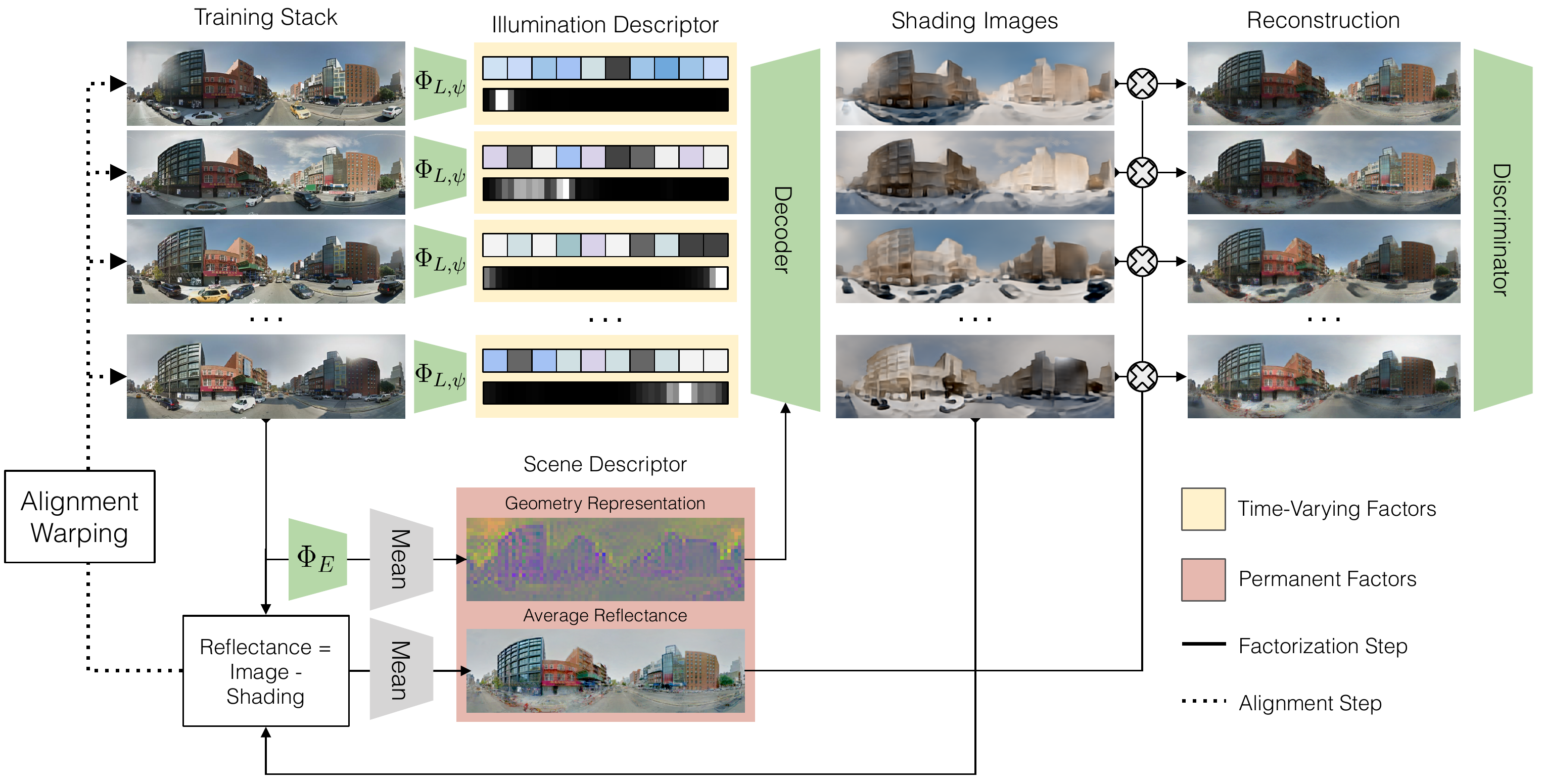}
    \vspace{-.3in}
    \caption{\textbf{Training with timelapses.} We train encoders to disentangle an assembled timelapse stack into two factors: \textit{illumination descriptors} that capture the time-varying aspects of each image, and a single \textit{scene descriptor} that captures the permanent elements of the entire timelapse stack, such as the scene geometry. We train a generator to transform the disentangled factors into shading and reflectance images from which we can reconstruct the original images. As indicated by the dotted pathways, we also simultaneously solve for the alignment of the individual frames in the input timelapse.}
    \label{fig:main}
\end{figure*}

Our model's latent factors are organized into two sets of descriptors, as shown in Figure~\ref{fig:main}: an \textit{illumination descriptor} represents temporally varying aspects of the scene and a \textit{scene descriptor} represents the permanent aspects.

\medskip
\noindent \textbf{Illumination Descriptor:}
Our illumination descriptor captures the factors of the world that encode temporal variation like lighting.
This descriptor is comprised of two disentangled sub-factors:

The \textit{lighting context} $L \in \mathbb{R}^{32}$ is a global latent feature that captures the overall ambient illumination properties, such as atmospheric conditions and cloud cover. Our lighting context encoder $\Phi_L$ encodes an image to this embedding.

The \textit{sun azimuth angle}, $\varphi$ is an explicit factor representing the horizontal position of the sun in a given panorama.
We model sun azimuth explicitly because, unlike illumination patterns, variations in sun azimuth have a simple geometric meaning, with a value in the range $[-\pi, \pi]$. Despite this simple parameterization, the effect of sun azimuth on a rendered scene is highly complex. Therefore an explicit azimuth factor allows our model to combine the factor's underlying mathematical simplicity with a network's ability to model complex behaviors.

Rather than regress to a scalar angle, we instead represent $\varphi$ internally as a discretized distribution over sun angle (with $k=40$ bins).
Inspired by prior work on illumination estimation~\cite{deepskymodel}, our azimuth encoder $\Phi_\varphi$ is a horizontally fully-convolutional network
that takes as input a panorama, and produces a 40-way softmax distribution $\varphi$,
where each bin corresponds to the probability that the sun azimuth is located in the bin's corresponding angular range. Note that given this discrete distribution over angles, we can differentiably compute a single scalar angle as the (circular) expectation of the distribution, $\bar\varphi$.
This predicted scalar sun angle is used by our decoder for normalizing sun position.

\medskip
\noindent \textbf{Scene Descriptor:}
Our scene descriptor captures the permanent structure of the world that is invariant to the temporally varying effects described above.
We also divide this descriptor into two disentangled sub-factors:

The \textit{geometry representation} is a spatial map of learned features that captures scene properties (e.g.\ surface normals and material properties) that are independent of illumination, but nonetheless are important to determining the rendering of a shading images. The fully convolutional encoder $\Phi_E$ outputs $E \in \mathbb{R}^{\frac{H}{8} \times \frac{W}{8} \times 16}$ where H and W are the resolution of a panorama.

The \textit{reflectance image} is an RGB estimate of the underlying scene albedo. In contrast to the shading image, we chose to not use an encoder-decoder to compute reflectance for two reasons: (1) neural networks can have difficulties preserving high-frequency textures that are important for visual quality and (2) it suffices to predict only one intrinsic image component because its complement component has a closed form solution based on the intrinsic images equation.

\medskip
\noindent \textbf{Decoder:}
Given a set of learned factors (sun azimuth angle $\bar\varphi$, lighting context $L$, and geometry factor $E$), our decoder $G$ is trained to generate an outdoor shading image. 
To facilitate training of $G$,
one insight is that it is easier to learn to synthesize shading images with a fixed sun azimuth angle than with all possible angles. Further, we can normalize a panorama by its predicted sun azimuth angle by simply rotating it by the negative of that angle (i.e., circular horizontal translation). Hence, our decoder operates as follows: (1) use the predicted sun azimuth angle $\bar\varphi$ to rotate the geometry factor image $E$ to a fixed sun angle, 
(2) decode the sun-normalized geometry image with lighting context $L$ to a shading image, and (3) rotate the result back to the original coordinate frame.

We use the Spatial Adaptive Instance Normalization (SPADE) generator of Park~\textit{et al.}~\cite{taesung} to model the complex interactions between geometry and illumination in our decoder $G$. The SPADE generator takes the lighting context $L$ as the network's noise input. We apply the insights from above and rotate the geometry representation $E$ by $-\bar\varphi$ before using it as the SPADE conditioning. 

While some prior works model shading with a grayscale image, such a model cannot capture real-world, colored illumination. 
Inspired by Sunkavalli~\textit{et al.\ }~\cite{ftlv}, we augment our decoder's gray-scale shading predictions with a bi-color assumption by additionally predicting two global color illuminants $c_1$ and $c_2$, corresponding to sunlight and skylight, and a per-pixel mixing weight $M$ that models how much each pixel is illuminated by the sun or sky.
For further details about the  
decoder architecture, please refer to the supplemental material.

\subsection{Training}
\label{sec:training}
Learning to factor single images without \textit{any} supervision is challenging---there is simply not enough information in a single image to disentangle scene factors from illumination factors. However, a GSV-TM stack depicts the same underlying permanent scene under diverse temporally varying illuminations, providing a useful training signal. Our training procedure, shown in Fig.~\ref{fig:main}, learns to disentangle factors \textit{within} a stack by separating the permanent geometry of the scene shared by all images in the stack from the varying lighting. The
trained model can be applied to a single image at test time.

Given a timelapse stack, we run our encoder on individual frames to get a stack of encoded geometry representations and illumination descriptors.
Because we assume the stack's geometry to be constant across time, we average the encoded geometry maps over 
the stack, resulting in a single shared geometry map, $\bar E$. From this shared geometry map, and the per-image illumination factors, our decoder produces a stack of shading and reflectance image pairs. As with geometry, we wish the scene's albedo to be constant across time. Accordingly, we impose a reflectance consistency loss $\mathcal{L}_\mathsf{RC}$ that computes the $L_1$ 
distance between pairs of reflectance images 
from different frames. This loss encourages the encoder-decoder network to remove temporal variation from the encoded permanent factors such that the 
reflectances are constant across a stack.

As demonstrated in the right half of Fig.~\ref{fig:main}, we average the stack's reflectance images across frames to get the stack's shared reflectance. The shared reflectance is recomposited with the shading image of each frame in the stack to reconstruct the original pixels of each input frame. These reconstructions are used to drive the learning process via image synthesis losses.

\subsection{Stack alignment}
\begin{figure*}[t]
    \centering
    \includegraphics[width=\linewidth]{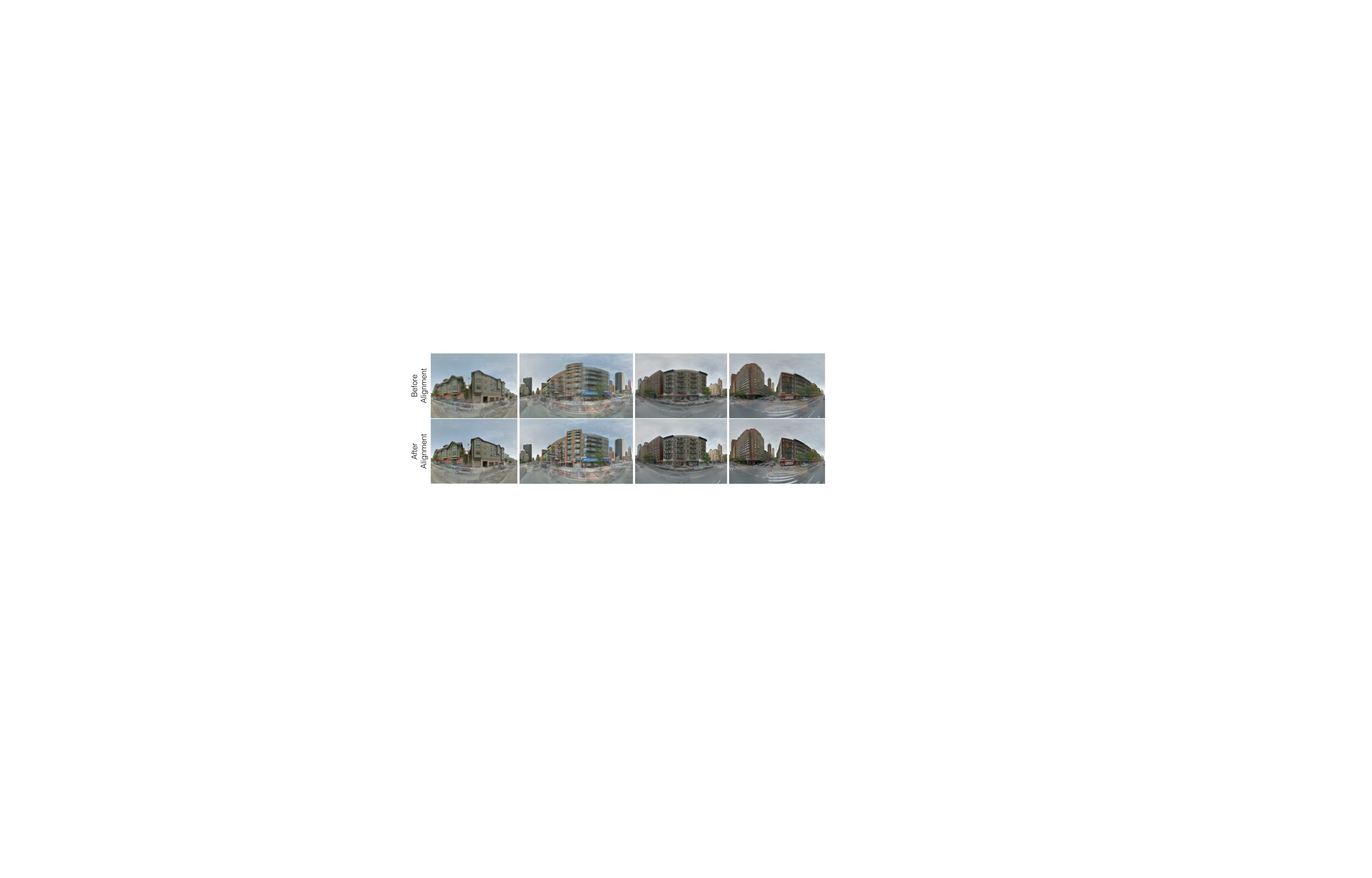}
    \vspace{-.3in}
    \caption{\textbf{Alignment results.} We show stack averages, cropped for emphasis, before and after our alignment process. Aligning the estimated permanent reflectances rather than the input images results in good alignment and therefore crisp stack averages.} 
    \label{fig:align-results}
\end{figure*}

Unlike traditional webcam data, our assembled GSV-TM timelapses do not come from stationary cameras. While each stack consists of nearby panoramas, they are not perfectly co-located and aligned. 
As shown in Fig.~\ref{fig:align-results}, the average of the stack reveals visible misalignment artifacts resulting from this parallax.

We could use 3D reconstruction methods as the basis for image alignment, but opted for a simpler 2D approach 
inspired by image congealing~\cite{congealing}, and compute 2D warps that best align the images in each stack. Given a raw stack of imperfectly aligned images, we define $\Theta$, an $8\times32$ grid of per-image control points initialized as the identity warp. The control points 
define a 2D spline used to differentiably warp each image within a stack to align with the rest.

To find the control points that best align images within a stack, we run gradient descent to minimize pixel alignment error. While one could 
use 
original image pixels to measure misalignment, 
we found that photometric differences across the stack due to varying lighting conditions 
led to poor alignments. Instead, we compute 
error on 
estimated \textit{reflectance} images by reusing our previously defined reflectance consistency loss, $\mathcal{L}_\mathsf{RC}$, to update alignment parameters. This approach is indicated by the dotted pathway in Fig.~\ref{fig:main}.
By jointly minimizing alignment and intrinsic image decomposition, we create a positive feedback loop---as timelapse alignment improves, factorization becomes easier and vice versa.

\subsection{Losses}
Our losses are optimized over alignment parameters $\Theta$, factorization encoders $\Phi_L$ $\Phi_\varphi$, and $\Phi_E$, and decoder $G$. We train a multi-scale patch discriminator~\cite{pix2pix2016,wang2018pix2pixHD} $D$ to ensure that the stack reconstructions with shared reflectances look realistic.

Our primary loss for learning the disentanglement is the reflectance consistency loss $\mathcal{L}_\mathsf{RC}$ described in Sec.~\ref{sec:training}. We include standard image generation losses on the reconstructed stack to ensure high quality synthesis results: a perceptual loss $\mathcal L_\mathsf{VGG}$~\cite{perceptualloss}, an adversarial loss $\mathcal{L}_\mathsf{GAN}$~\cite{biggan}, and a feature matching loss $\mathcal{L}_\mathsf{FM}$~\cite{wang2018pix2pixHD}. 
Finally, because intrinsic images have a fundamental color ambiguity, we also include a white light penalty, $\mathcal{L}_\mathsf{WL}$
that biases our encoder-decoder towards white-balanced reflectance outputs. Our overall objective function is:
\begin{equation}
\min_\Theta \max_D \min_{G, \Phi_L, \Phi_\varphi, \Phi_E} \mathcal{L}_\mathsf{RC} + \mathcal{L}_\mathsf{Gen} + \mathcal{L}_\mathsf{GAN}
\end{equation}
where $\mathcal{L}_\mathsf{Gen}$ is a weighted sum of $\mathcal{L}_\mathsf{FM},\mathcal{L}_\mathsf{WL},\mathcal{L}_\mathsf{VGG}$ that measures the generative quality of the reconstructed images.
We include additional descriptions, alignment results, insights, and analysis for reproducibility in the supplemental material.

\section{Experiments}
\label{sec:evaluation}
We evaluate our factorization method in two ways: 1) we compare to intrinsic image decomposition baselines in the single-scene setting, and 2) we apply our method to the task of transferring illumination descriptors across different scenes, a new capability enabled by our disentanglement. In both cases, we measure success by the quality of reconstructed images derived from swapping their disentangled factors with ones borrowed from other images as in~\cite{zhou2015intrinsic}. 

\medskip
\noindent \textbf{Data.}
At test time, our network can take as input either an assembled timelapse stack or a single panorama. In order to align test-time stacks like those shown in Fig.~\ref{fig:align-results}, we estimate spline parameters by computing a gradient for alignment only, while keeping the weights of the factorization part of the network frozen. Below, we present results for stack as well as single-image inputs.

In particular, we 
show single-image test-time results on 
GSV imagery
from cities never seen during training, such as Paris, 
as well as images from the Outdoor Laval HDR dataset~\cite{deepskymodel}. This dataset contains HDR panoramas of outdoor scenes that are tonemapped to sRGB to match GSV. We use this data to compare to existing sRGB intrinsic image methods 
and to test generalization from GSV to a different domain of panoramas.

\medskip
\noindent \textbf{Baselines.}
Given the novelty of our problem, we perform model ablations to measure the individual benefits of various components. All ablated models are trained with the same losses and number of iterations as our full method. We report results on the following ablations:

\begin{itemize}[leftmargin=*]
\item\textbf{Mono-color shading}: We ablate the bi-color shading by training our model with a mono-color assumption similar to that of Li and Snavely~\cite{BigTimeLi18}.

\item\textbf{w/o alignment training}: Trained without the alignment feedback loop.

\item\textbf{w/ unaligned test stacks}: Uses unaligned test stacks to measure the effect of ablating alignment at training (above) vs.\ at both training and test time.

\item\textbf{w/o azimuth encoder}: Our model trained without an azimuth encoder nor normalizing for sun position.
\end{itemize}
\noindent
Additionally, we consider the following baselines:
\begin{itemize}[leftmargin=*]
\item\textbf{Pixel nearest neighbor}: 
Given a target image, we find the pixel-wise nearest neighbor in its aligned stack and report the error resulting from using that image as our synthesized result. 

\item\textbf{Weiss's MLE Intrinsics}~\cite{weiss2001intrinsics}: use handcrafted priors on gradients extracted from image sequences.

\item\textbf{Zhou~\textit{et al.}}~\cite{zhou2015intrinsic}: learn to mimic human judgments of relative reflectance.

\item\textbf{Li and Snavely's BigTime}~\cite{BigTimeLi18} learn shading priors from image sequences.
\end{itemize}

\subsection{Within-Scene Decomposition}
\label{sec:intrinsicimage}
\begin{figure*}[t]
    \centering
    \includegraphics[width=\linewidth]{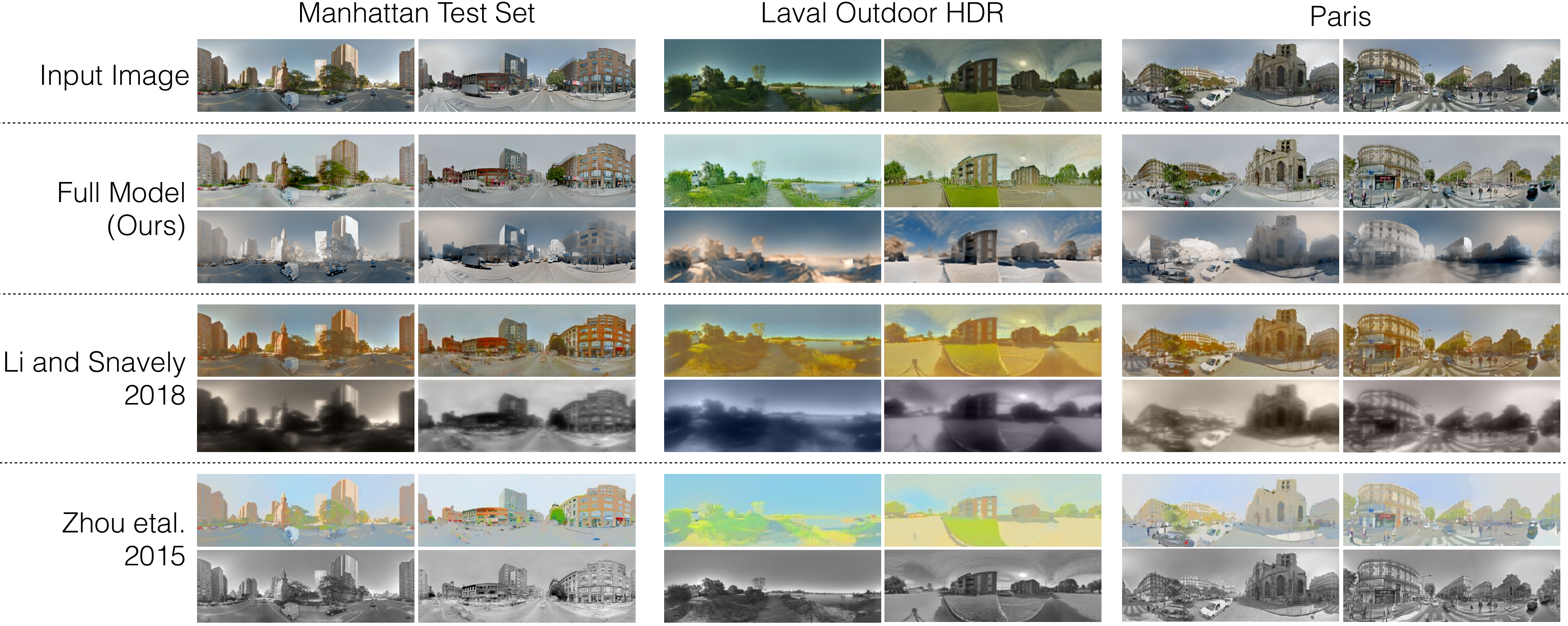}
    \vspace{-.3in}
    \caption{\textbf{Qualitative results on an intrinsic image decomposition task.}
    We compare single-image decompositions of our method with Li and Snavely~\cite{BigTimeLi18} and Zhou \textit{et al.}~\cite{zhou2015intrinsic}. 
    Compared to the baselines, our reflectance images do not have residual shadows.
    Our method, trained on NYC, generalizes at test-time to Laval Outdoor HDR Panoramas~\cite{deepskymodel} as well as to GSV imagery from Paris.}
    \label{fig:iii_compare}
\end{figure*}

\begin{figure*}[t!]
    \centering
    \includegraphics[width=\linewidth]{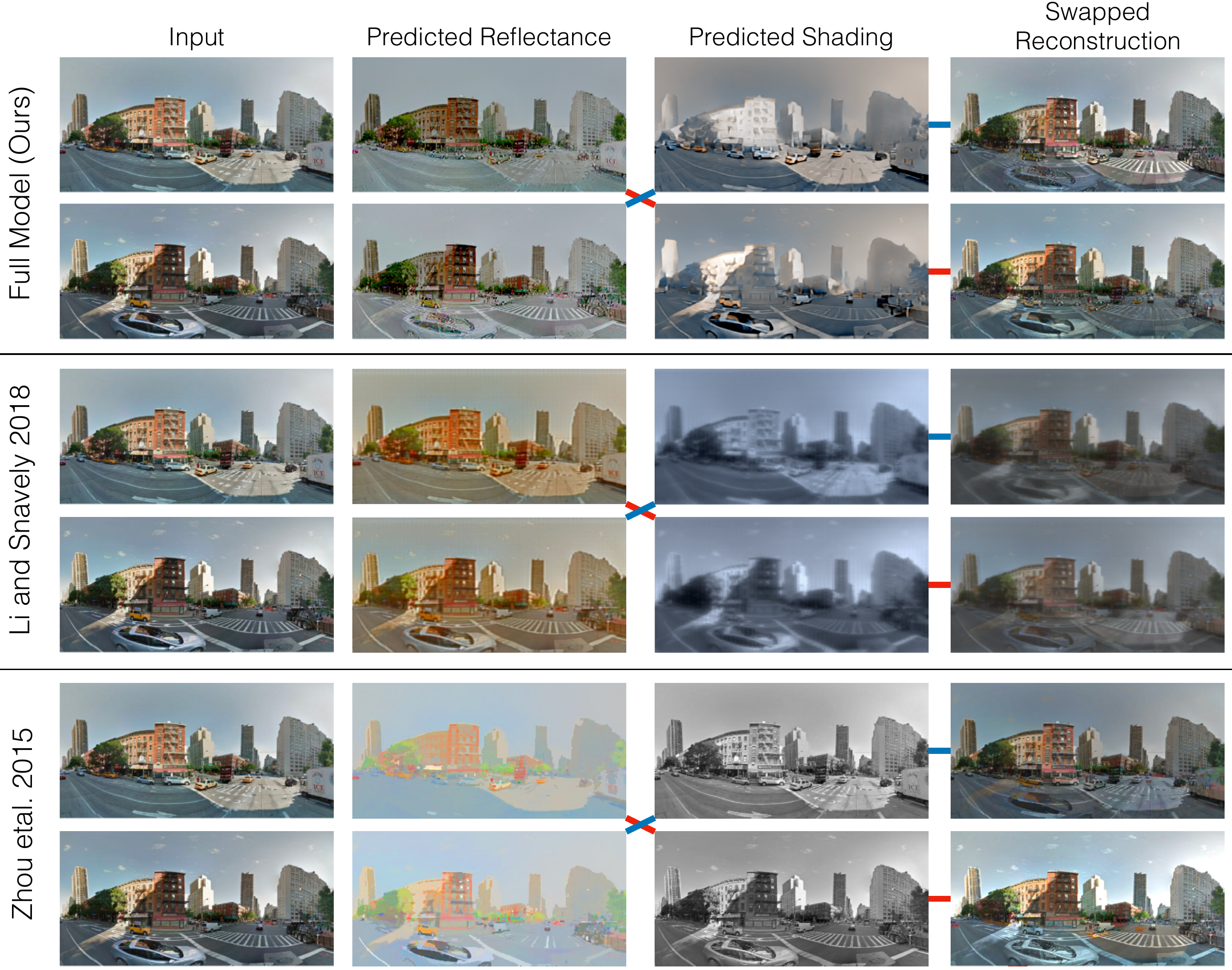}
    \vspace{-.3in}
    \caption{\textbf{Transferring illumination within a scene.} Given a pair of images of the same scene under different illuminations~(\textit{left}), we disentangle the permanent and varying factors and decode their reflectance and shading~(\textit{middle}). To test the permanency of the estimated reflectance for the depicted scene, we swap reflectances within the pair and combine them with the estimated shading to reconstruct the original images~(\textit{right}). Red and blue paths connect the components used to reconstruct each image. Our method produces a reflectance, clean of any lighting, which can be safely swapped between captures of the same scene and still result in good reconstructions.}
    \label{fig:scene_swap}
\end{figure*}

Intrinsic image methods aim to decompose an image into shading and reflectance. The quality of a decomposition is measured by its ability to separate illumination effects, like cast shadows, from permanent properties such as albedo.
In Fig.~\ref{fig:iii_compare}, we show reflectance and shading computed from a single image using our method and the two deep learning baselines. Both BigTime and Zhou~\textit{et al.} fail to remove cast shadows, as seen by residual shadows encoded in their reflectance. Unlike Zhou~{\textit{et al.}}, our method produces shading images that are piecewise smooth, as expected for planar surfaces like building facades. BigTime struggles in outdoor settings because their single global illuminant cannot predict multiple illumination colors. Finally, both baselines incorrectly encode blue sky pixels as reflectance despite the fact that sky color is a temporal property. To further illustrate the advantages of our method over these baselines, Fig.~\ref{fig:scene_swap} shows the results of relighting pairs of images of the same scene by swapping reflectances within the pair. Unlike the baselines, our clean reflectance image allows us to relight the scene successfully.

\medskip
\noindent \textbf{Scene consistency verification.}
Since MLE Intrinsics~\cite{weiss2001intrinsics} only works on timelapse stacks of single scenes, we devise a way to quantitatively compare to their method. We split our aligned test stacks to two smaller substacks of 4 images each. For each substack, each method predicts 
a single reflectance image and four shading images. Since both substacks capture the same underlying scene, the predicted reflectances should be consistent across the two. As in the case of single images (Fig.~\ref{fig:iii_compare}), we can test the consistency of the predicted reflectance for the depicted scene by swapping the predicted reflectance images between the two substacks and reconstructing the four input images in each substack from their shading and \textit{swapped} reflectance images. We refer to this experiment as \textit{scene consistency verification} because the reconstruction error is minimized when the predicted reflectances are identical for the two substacks.

\setlength{\tabcolsep}{4pt}
\begin{SCtable}[\sidecaptionrelwidth][t]
\begin{minipage}[t!]{0.52\linewidth}
\small
\centering 
\resizebox{\columnwidth}{!}{
\begin{tabular}{lcc}
\toprule
Model & Consistency & Completion \\
\midrule
Full model (ours)                    & \textbf{0.071} & \textbf{0.196} \\
Mono-color shading                   & 0.077 & 0.215 \\
w/o alignment training                        & 0.082 & \textbf{0.201} \\
w/ unaligned test stacks & 0.090 & 0.210 \\
w/o azimuth encoder                  &\textbf{0.072} & 0.240 \\
\midrule
Pixel nearest neighbors              & 0.274 & 0.278 \\
MLE Intrinsic~\cite{weiss2001intrinsics} & 0.114 & --- \\
BigTime~\cite{BigTimeLi18}               & 0.180 & --- \\
Zhou~\textit{et al.}~\cite{zhou2015intrinsic}& 0.217 & --- \\
\hline
\end{tabular}
}
\end{minipage}
\hfill
\begin{minipage}[t!]{0.48\linewidth}
\caption{\textbf{Relighting results.} We define two image reconstruction tasks for evaluation.  \textit{Scene consistency verification} evaluates whether the estimated reflectance is consistent across multiple captures of a single scene. \textit{Space-time completion} evaluates the ability to transfer illumination across different scenes. We report MSE reconstruction error. Lower is better.
}
\label{table:swap}
\end{minipage}
\end{SCtable}
\setlength{\tabcolsep}{4pt}

We report the mean squared reconstruction error (MSE) between the input stack and the swap reconstructions in Table~\ref{table:swap}. 
Our method outperforms 
the three baselines 
at image reconstruction 
in this setting. 
We speculate that prior methods are hindered by their reliance on hand-defined shading priors and limited training data.
In contrast, our massive dataset provides enough supervision for learning a good decomposition without shading priors. Interestingly, ablating the azimuth encoder does not degrade performance on this task, suggesting that a simpler setup is 
sufficient for within-scene illumination transfer. 

\subsection{Cross-Scene Factorization}

Unlike 
intrinsic images methods, our factorization 
allows us to transfer illumination descriptors \textit{across} scenes.
Using our disentangled factors,
we can synthesize a given scene under completely new lighting conditions, borrowed from a \textit{different} location. For the purpose of evaluating the success of this cross-scene relighting process, we devise a way to compare the novel synthesis to ground truth. Namely, because illumination changes relatively slowly, we assume that images captured within 5 minutes across the city have the same illumination descriptor. 
Hence, we 
can relight a given scene, $A$, captured at time $T_1$ using illumination descriptors transferred from a different location, $B$, captured at time $T_2$. We then compare the resulting synthetic image of scene $A$ at time $T_2$ to ground truth captures of scene $A$ captured at a time close to $T_2$.

We name this task \textit{space-time matrix-completion}. A row in the matrix represents a unique point in ``space'' and a column represents a unique point in ``time''. A single panorama represents an entry in this matrix at the row corresponding to its depicted scene and column corresponding to its capture time. We can withhold entries in the matrix and reconstruct 
them by combining a scene descriptor derived from images in the same row, with an illumination descriptor extracted from a different scene from the same column. Table~\ref{table:swap} shows the reconstruction MSE for each ablation between  held-out and reconstructed views. Our full model and the \emph{w/o alignment training} ablation show significant improvements over other ablations. 

While alignment training does not significantly affect the performance of our model on this task (\textit{w/o alignment training}), its performance
degrades significantly on unaligned stacks (\textit{w/ unaligned test stacks}). This indicates that alignment may be optional during training but is crucial for reconstruction. Additionally, unlike with the substack swap task,
explicitly representating sun azimuth improves transferability of lighting descriptors across scenes.

\section{Applications}

We now present applications where we synthetically modify a panorama. These applications are uniquely enabled by our intrinsic factorization that disentangles time-varying effects from the permanent scene properties.

\medskip
\noindent \textbf{Changing sun position.}
\begin{figure*}[t]
    \centering
    \includegraphics[width=\linewidth]{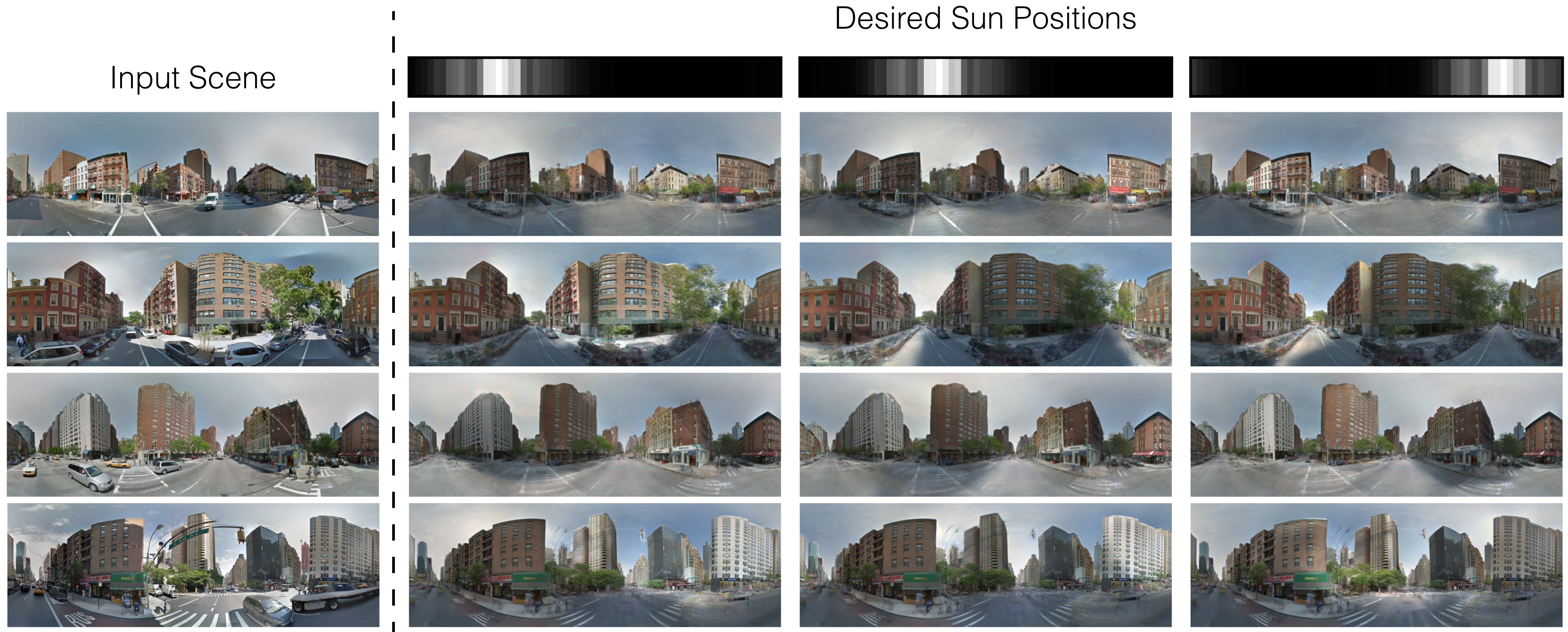}
    \vspace{-.3in}
    \caption{\textbf{Manipulating sun position.} We can specify the sun position for an input scene and relight it realistically. Please see the supplemental video for full animations.}
    \label{fig:sun_rot}
\end{figure*}

Our model disentangles sun azimuth angle from scene and lighting context factors. Once  a scene is factorized, we can visualize what a scene looks like when the sun angle is changed. Fig.~\ref{fig:sun_rot} shows examples of test scenes synthesized with new sun azimuth angles. Note that cast shadows and illumination on building faces change realistically with the rotation.

\medskip
\noindent \textbf{Relighting a \textit{novel} scene.}
\begin{figure*}[t]
    \centering
    \includegraphics[width=\linewidth]{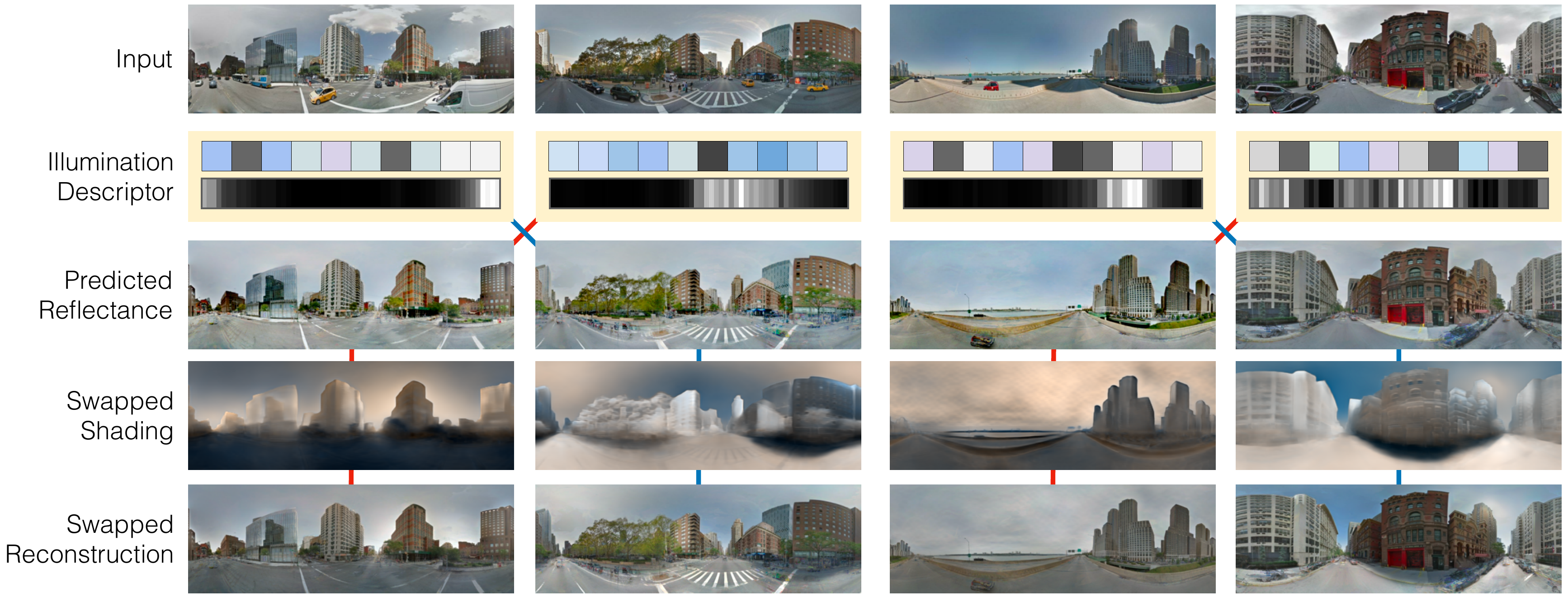} 
    \vspace{-.3in}
    \caption{\textbf{Changing sky illumination.} We can relight \textit{novel} scenes by transferring the disentangled time-varying factors from one scene to another. Here we swap the illumination descriptors of a pair of input scenes to visualize what each scene might look like under a new illumination. The red and blue paths indicate the the components used to reconstruct each relit scene.} 
    \label{fig:swap-ill}
\end{figure*}

Our lighting context encodes the stylistic quality of illumination. As shown in Fig.~\ref{fig:swap-ill}, we can transfer the whole illumination descriptor, including sun azimuth, from one panorama to another with a new scene geometry. Results for transferring \textit{only} lighting context can be found in the supplemental material. The supplemental material also demonstrates relighting a spatial sequence of panoramas from different times to a fixed illumination, thus producing a virtual drive through Manhattan.

\medskip
\noindent \textbf{Editing scene geometry.}
While shading and azimuth capture the essence of time, the scene descriptor encodes structures. 
By copy-pasting regions of the scene descriptors, we can transplant the buildings into new panoramas and relight them to match the scene. Please see the supplementary for results.

\section{Discussion}
We proposed a novel source of large-scale timelapse data from historical Street View data, and a learning-based method for factorizing temporal and permanent variations across imagery covering an entire city. 
Our learned factorization outperforms state-of-the-art intrinsic images methods, and enables cross-scene style transfer via manipulating our learned factors.

Our method has a few 
limitations. First, the scene descriptor learns to encode transient objects like cars. While moving objects are temporal effects, the network chooses to encode them in the scene descriptor, resulting in wispy cars appearing in the generator output. Second, high-frequency details such as cast shadows from tree branches are difficult to synthesize. 
Third, when the alignment module fails, the 
shared reflectance of a stack will appear blurry. Please see the supplemental material for examples of failure cases. Finally, when our permanence assumptions fail to hold---for instance when buildings are repainted or rebuilt---our assumption that the scene descriptor is constant across time is violated.

Despite these limitations, our work points towards a new approach to modeling and synthesizing the space of outdoor scenes, wherein we can learn to separate factors that persist at different time scales. An intriguing direction for future work is to expand to a richer range of timescales, for instance modeling transient effects (moving people, cars, etc), effects with annual cycles (e.g., seasons), long-term changes like weathering, etc..

\smallskip
{\small
\noindent\textbf{Acknowledgements:} We would like to thank Richard Tucker, Richard Bowen, Ameesh Makadia, and Vincent Sitzmann for insightful discussions. We would also like to thank Angjoo Kanazawa and Tim Brooks for their help with preparing the manuscript. This work is supported, in part, by NSF grant IIS-1633310.}

\clearpage
\bibliographystyle{splncs04}
\bibliography{refs}

\clearpage
\setcounter{section}{0}
\renewcommand\thesection{\Alph{section}}

\section{Gamma Correction:}
The intrinsic images formula factorize an image \textbf{I} into reflectance \textbf{R} and shading \textbf{S} components.

\begin{equation}
    I = R \times S
\label{eq:intrinsic_image}
\end{equation}

Where I is assumed to be a linear RGB intensity scale. Ideally these linear RGB images should come from high-dynamic range (HDR) exposures because they capture the full spectrum of illumination which are factored into the shading component. However many images, including Google Street View panoramas, are tone-mapped and encoded in the standard RGB (sRGB) colorspace. The tone-mapping procedure involves clipping exposures and gamma correction such that values above the clip are visualized as over-exposed pixels. The gamma correction relates sRGB and linear RGB pixels:
\begin{equation}
    I_\mathsf{sRGB} = A I^\gamma
\end{equation}
Where $A = 1$ and $\gamma = \frac{1}{2.2}$ typically.

Because tone-mapping irrecoverably loses information about the original lighting, we simply assume the overexposed linear RGB pixels recovered from reversing the gamma correction sRGB are the true brightness. In practice this suffices for intrinsic images as seen by Weiss's MLE Intrinsics~\cite{weiss2001intrinsics}, Zhou~\etal~\cite{zhou2015intrinsic}, and Li and Snavely's BigTime~\cite{BigTimeLi18} which do gamma ``uncorrection" by scaling the sRGB image to take values between [0, 1] and then raising the pixels to the power $\frac{1}{\gamma}$. 

However one common trick with manipulating intrinsic images is to work in log-space. More specifically one can take the intrinsic image formula (Eq.~\ref{eq:intrinsic_image}) and apply the log function to linearize the relationship between log-shading and log-reflectance.

Our observation is that by replacing linear RGB input with the sRGB, the intrinsic image formula becomes:

\begin{equation}
    \frac{1}{\gamma}\log(I_\mathsf{sRGB}) = \log(R) + \log(S)
\label{eq:intrinsic_image}
\end{equation}

In log-space, the log-reflectance and log-shading components computed with sRGB and linear RGB are related by a scale $\frac{1}{\gamma}$. In our system, the SPADE decoder $G$ is a learned convolutional neural network that outputs log-shading. Therefore the scale factor $\frac{1}{\gamma}$ can be trivially learned by the convolutional filter weights of the last layer. This has been confirmed as we trained a model using linear RGB and $\frac{1}{\gamma}=2.2$ and achieved comparable results to ones trained with sRGB. All results shown in submission and main paper have been trained with sRGB inputs \textbf{without} extra handling for gamma correction.

\begin{figure}[t]
    \begin{center}
    \includegraphics[width=0.95\linewidth]{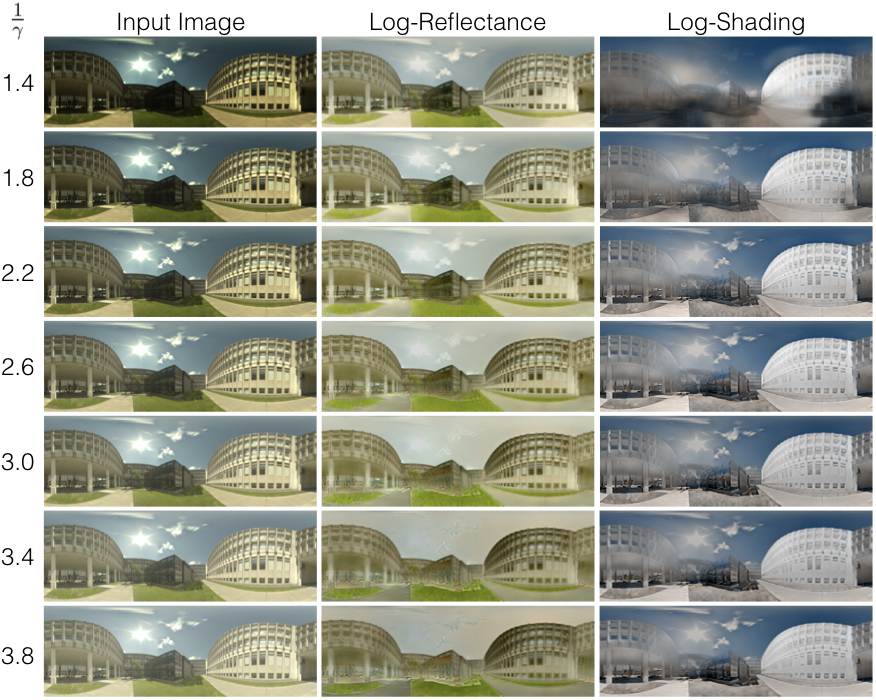}
    \end{center}
    \caption{We show different gamma curves applied to an original HDR panorama capture from the Laval Outdoor HDR dataset~\cite{Hold-Geoffroy_2019_CVPR}.
    The standard gamma correction is computed with $\frac{1}{\gamma}=2.2$. For each different value of gamma, we show our model factorizing the resulting gamma corrected image. As shown by the degradation of performance at $\frac{1}{\gamma}=1.4$, our model is sensitive to decomposing images with unconventional gamma correction.}
    \label{fig:gamma-correction}
\end{figure}

One potential problem comes from attempting to decompose images with unconventional gamma curves. We show in
Fig.~\ref{fig:gamma-correction} an HDR capture from the Laval Outdoor HDR dataset~\cite{deepskymodel} that has been tone-mapped with various gamma values. While our model performs well near the typical parameter $\gamma=\frac{1}{2.2}$, the decomposition degrades especially at $\gamma=\frac{1}{1.4}$. This is an out-of-distribution training problem as most images trained by the model have been tonemapped with the standard $\gamma=\frac{1}{2.2}$. While training with linear RGB images would resolve any ambiguities between different gamma curves, the fundamental problem is that the gamma parameter for arbitrary sRGB images is unknown and assuming $\gamma=\frac{1}{2.2}$ is a best approximation for most images, in which case the correction is learned by the decoder.

\section{Method}
\begin{figure}[t]
    \begin{center}
    \includegraphics[width=0.5\textwidth]{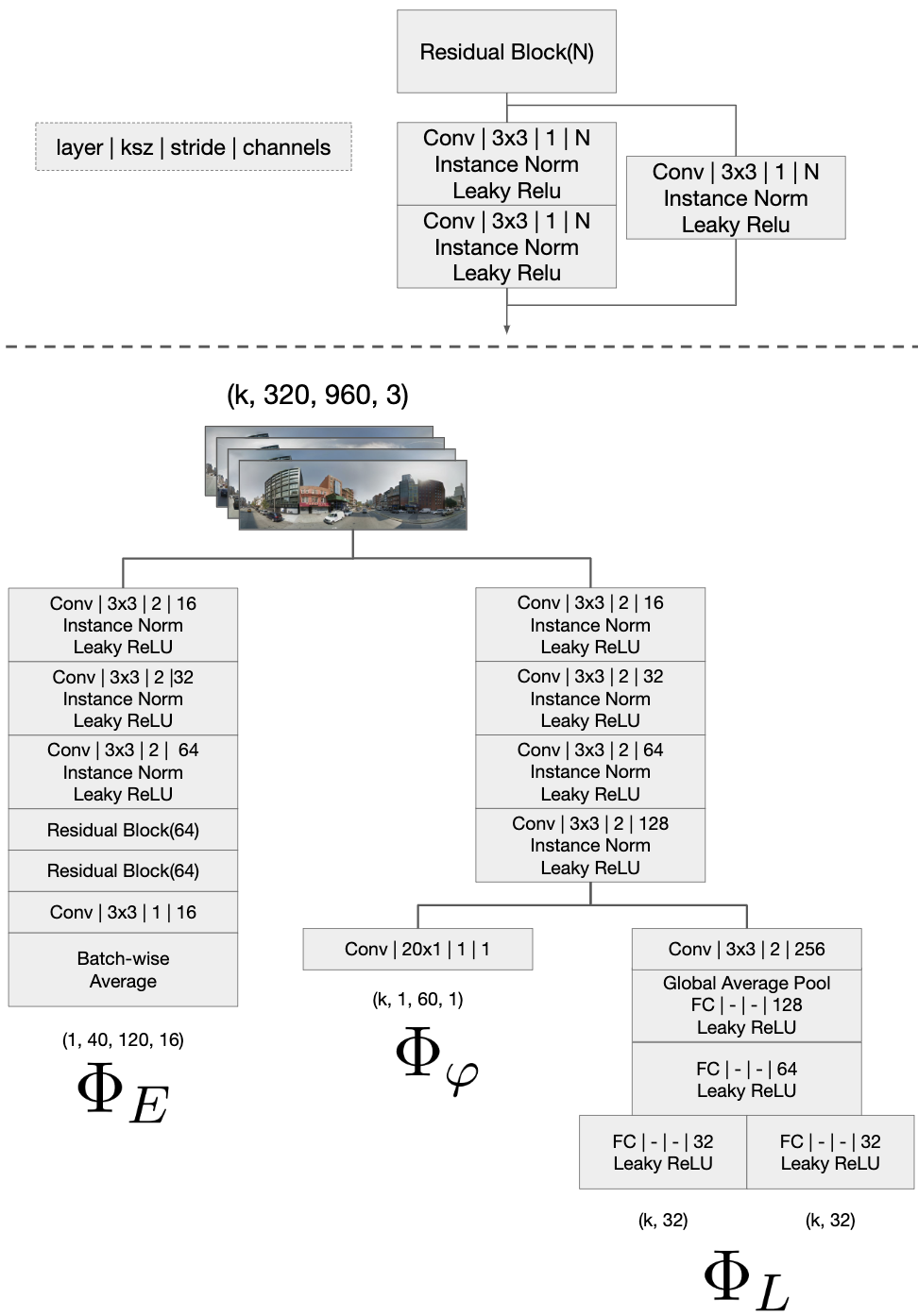}
    \end{center}
    \caption{Our three encoder architectures. The two encoders that output our illumination descriptor, azimuth and lighting context, share the first few convolutional layers before diverging into a horizontal fully-convolutional encoder and a variational encoder.}
    \label{fig:encoders}
\end{figure}
We show in Fig.~\ref{fig:encoders} a diagram of our encoder $\Phi_E, \Phi_\varphi, \Phi_L$. The two illumination descriptor encoders have a shared encoder weights as they are designed to extract transient effects.

\subsection{Sun Azimuth Encoder}
\label{sec:sa_encoder}
\noindent
\textbf{Correction:} In the main paper we mistakenly say that $\varphi$ is a 40-way classification problem. This is a typo and we intended to say that $\varphi$ is a \textit{60-way classification problem}.

Recall that as a pre-process, all training panoramas are oriented with a consistent heading---the center column faces the same 3D world direction, and the horizon corresponds to the center scanline.
In this format, cyclic horizontal translations of the panorama correspond to changing the heading in world coordinates. For example, if a north-facing panorama is shifted by half the length of the image, the resulting panorama will be the same scene, but facing south.

Given the shift-equivariant property of 
standard convolutions~\cite{MCSIA}, and the equivalence between shifts and rotations for horizon-levelled panoramas, standard convolution operations on panoramas are 3D-yaw-rotation-equivariant. This implies that sun azimuth angle can only be estimated \textit{relative} to the heading of the panoramas, as opposed to predicting an absolute cardinal direction of the sun.

In order to preserve equivariance we must use a fully convolutional encoder network as any global pooling layers would affect the spatial relationship between input and output and result in a network that is invariant (rather than equivariant) to rotations.

The output of the fully convolutional sun azimuth classification network is of dimension $12 \times 60$ where 60 is the number of angle classes and the first dimension can be pooled over as it is does not encode angle information to obtain a $1 \times 60$ distribution over angles. To this end, we learn a $12 \times 1$ learned vertical pooling layer rather than use a fixed max or average pooling operation. The full encoder is thus a horizontally fully-convolutional network, $\Phi_\varphi$.

\subsection{Panoramic Spatial Transformer}
\begin{figure}[t]
    \begin{center}
    \includegraphics[width=1.0\linewidth]{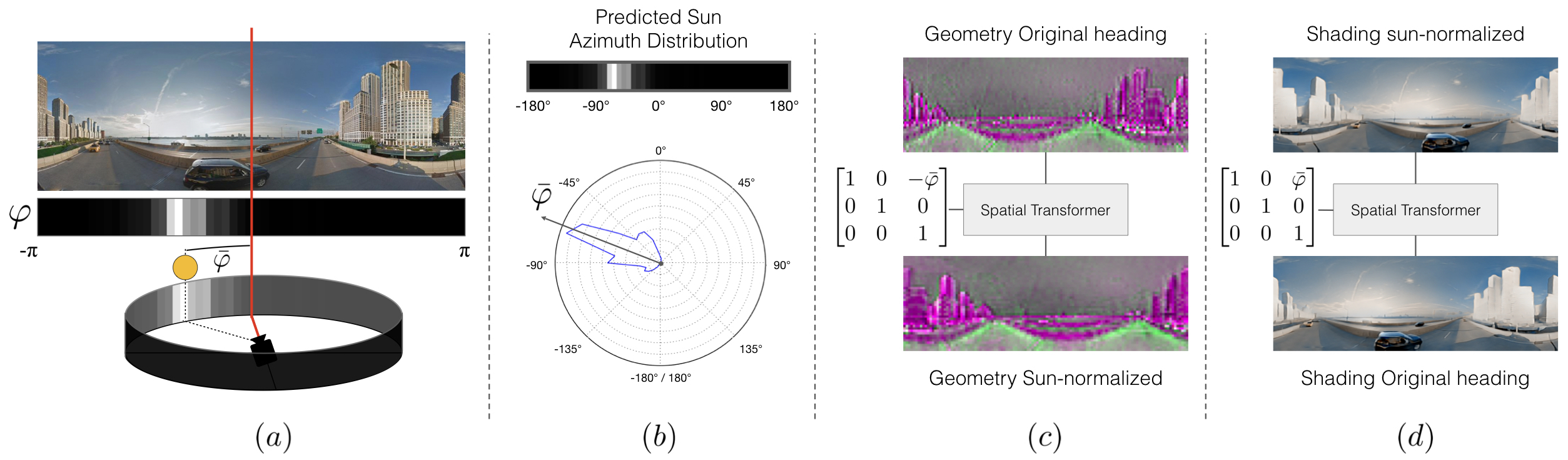}
    \end{center}
    \caption{Our \textit{Panoramic Spatial Transformer} is a novel representational technique for encoding and decoding 1D-rotations (formally SO(2) groups). In (a) we show an input panorama and the output $\varphi$ of our horizontally fully convolutional encoder $\Phi_\varphi$. The red line denotes the canonical heading which defines the zero-heading of $\varphi$'s coordinate system. We assume the canonical heading to always be the center of the panorama. In (b) we visualize the top-down view of the distribution as well as the definition of $\bar\varphi$ which is the circular average of the azimuth distribution $\varphi$. In (c) we take the geometry representation $E$ and apply a horizontal spatial transformer parameterized by $-\varphi$ which rotates $E$ to the orientation where the sun's position is at the canonical heading. Finally in (d) we show the generated shading with the sun-normalized geometry. To restore the original heading of the image we apply a spatial transformer parameterized by $\bar\varphi$ to get the shading of the input panorama.}
    \label{fig:pano_stl}
\end{figure}

We wish to rotate the geometry representation by the predicted sun azimuth such that the sun is always at the same angle (\textit{e.g.\ } head on). However, the softmax distribution, $\varphi$, over the 60 sun azimuth classification buckets spanning $[-\pi, \pi]$ as is described above in Section~\ref{sec:sa_encoder} is not differentiable. Luckily, circular angles are ordinal rather than categorical in nature. Therefore, instead of taking a softmax we can compute $\bar\varphi$, the circular average which is the expected value of this distribution, in a differentiable way:

\begin{equation}
\bar\varphi = \arctan\bigg( \frac{\mathbf{E}_{\alpha' \sim \varphi}[\sin(\alpha')]}{\mathbf{E}_{\alpha' \sim \varphi}[\cos(\alpha')]} \bigg) \label{eq:circ_ave}
\end{equation}

We note that the panoramic rotation operator describes a more general layer that excels at discovering the effects of 1D-rotations. In Fig.~\ref{fig:pano_stl}(a) we show an input panorama its azimuth distribution produced by $\Phi_\varphi$. We define the coordinate system of $\varphi$ relative to a consistent direction across all training examples. This direction is called the \textit{canonical heading} and is indicate by the red line. The canonical heading defines the $0^\circ$ coordinate with the negative angles to $-\pi$ represent orientations to the left of the canonical heading. Fig.~\ref{fig:pano_stl}(b) shows a polar plot of the distribution of sun azimuth orientations. $\bar\varphi$, visualized as the arrow, is about $-70^\circ$ with respect to the canonical heading. 

We can use $-\bar\varphi$ to parameterize a 1D spatial transformer that rotates the coordinate system to one that is invariant to sun-position as shown in Fig.~\ref{fig:pano_stl}(c). This orientation is called sun-normalization because after rotating by $-\bar\varphi$, all geometry representations are oriented with a consistent sun azimuth at $0^\circ$. Our insight is that the decoder's job is simplified if the sun is always in the same position. In Fig.~\ref{fig:pano_stl}(d) we show a decoded shading that is sun-normalized. To restore the original heading we use the 1D spatial transformer again parameterized by $\bar\varphi$.

By normalizing out sun azimuth, our downstream networks become invariant to sun azimuth, thereby teasing out a disentanglement between the orientation of the sun ($\bar\varphi$) and orientation of buildings in the world. Note that this must be used in conjunction with modifications for breaking rotation equivariance described in the next section. 

\subsection{Generator Architecture:}
\begin{figure*}[t]
    \begin{center}
    \includegraphics[width=1.0\linewidth]{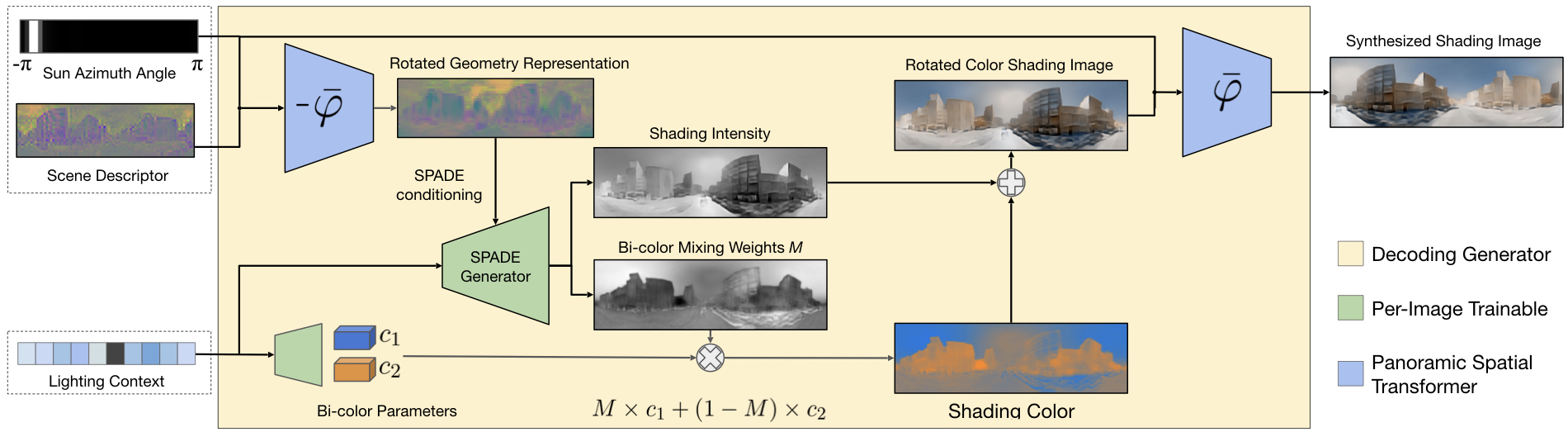}
    \end{center}
    \caption{The decoding generator consists of a Panoramic Spatial Transformer, a SPADE Residual Generator and bi-color shading estimation.}
    \label{fig:generator}
\end{figure*}

Our generator is derived from Park~\textit{et al.\ }~\cite{taesung} SPADE residual blocks. More specifically we use noise sampled from the parameters estimated by variational lighting context encoder and condition using SPADE residual blocks described in \cite{taesung} supplemental. The SPADE residual blocks are conditioned with the azimuth-rotated geometry code described above.

\medskip
\noindent
\textbf{Breaking rotation equivariance.}
Even though we have rotated our geometry code, the equivariance property described in Sec.~\ref{sec:sa_encoder}, means that the decoder $G$ will produce the exact same activations as the unrotated version except shifted by the same amount as the original rotation. Therefore we want to break the rotational equivariance property to allow the network to learn to synthesize different activations for different rotations of the same geometry representations.

\begin{figure}[t]
    \begin{center}
    \includegraphics[width=0.8\textwidth]{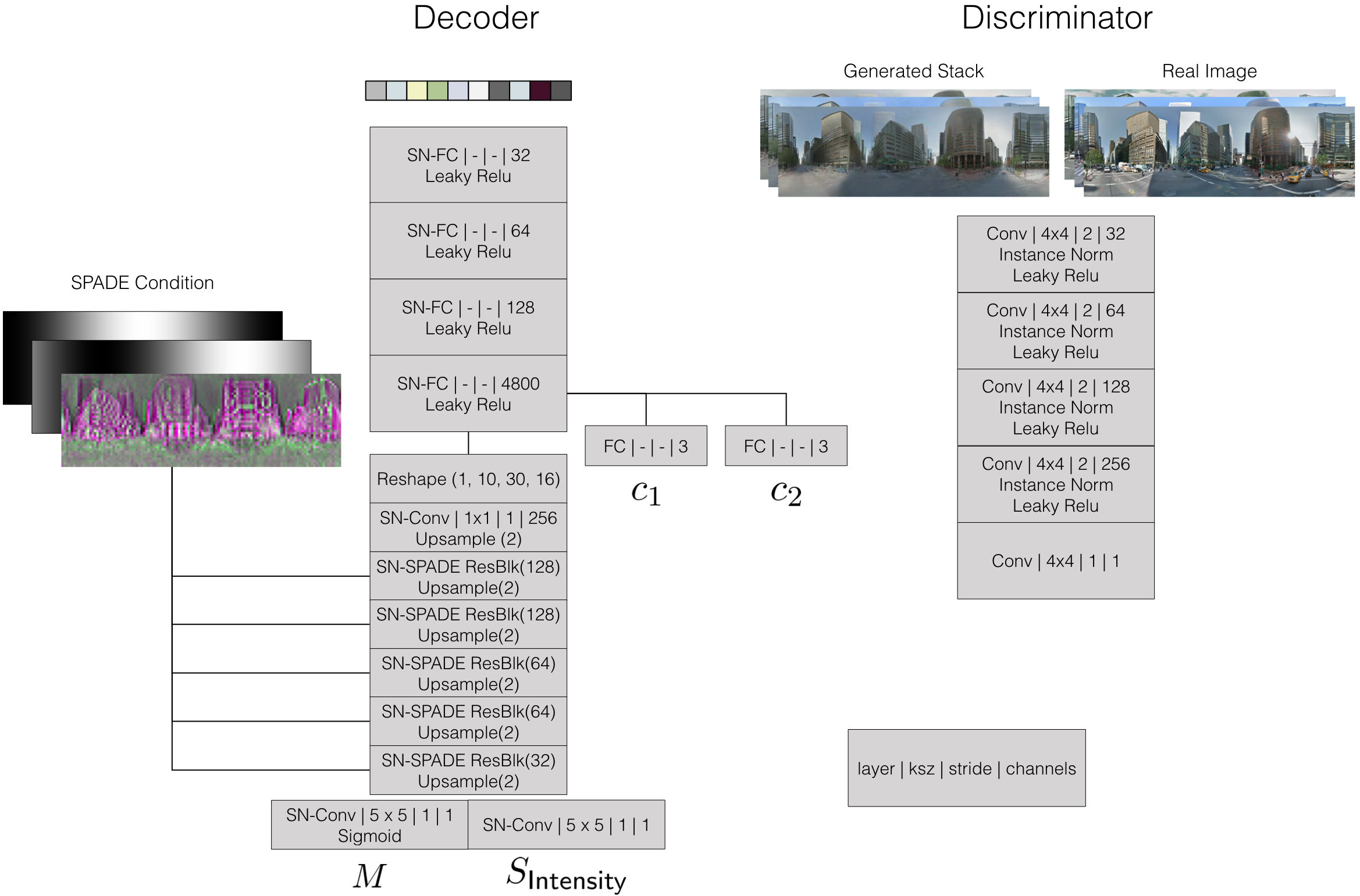}
    \end{center}
    \caption{We show our decoder and discriminator's setup. On left, our decoder uses spectral normalization~\cite{miyato2018spectral} fully-connected layers to decode global illuminants $c_1$ and $c_2$. SPADE Residual Blocks~\cite{taesung} are used to generate the Mask and Shading Intensity. Please refer to~\cite{taesung} for more information about SPADE Residual Blocks. On right we show our PatchGAN discriminator which outputs real and fake logits.}
    \label{fig:decoder-disc}
\end{figure}

To do this, we take one period of a sine and cosine signal from $[-\pi, \pi]$, sample 960 times (or the width of our panoramas), and tile the sampled signal vertically into an image that matches the height of our panoramas. The two images, shown in Fig.~\ref{fig:decoder-disc}(SPADE Condition), form a $(320, 960, 2)$ tensor which we resize and concatenate appropriately to the geometry code when its fed into the SPADE conditioning module. While breaking rotational equivariance is important, the implementation of the cosine and sine image can be effectively ignored because it is packaged and self-contained within our SPADE Generator which is \textbf{NO LONGER} rotationally equivariant. While we have described an intuitive way to break rotational equivariance, there exists other ways like a learned constant that is concatenated in place of the cosine/sine images.

We show in Fig.~\ref{fig:generator} the full generator diagram from factors to log-shading output. For information about the intermediate stages that form the bi-color module, please see refer to Sec.~\ref{sec:bi-color-supp}.

\medskip
\noindent
\textbf{Misc.}
We modify parts of SPADE to account for its new data and usage. We use panorama padding which involves padding with image content from the opposite side of the tensor in the width dimension. To ensure that gradients flow well through the SPADE's condition input, we use average-pooling for downsampling and nearest-neighbors for upsampling. Spectral normalization~\cite{miyato2018spectral} is used in both the main pathway and the convolutions embedded in the SPADE normalization layers. For explicit details about implementation please see Fig.~\ref{fig:decoder-disc}.

\begin{figure*}[t]
    \begin{center}
    \includegraphics[width=0.95\linewidth]{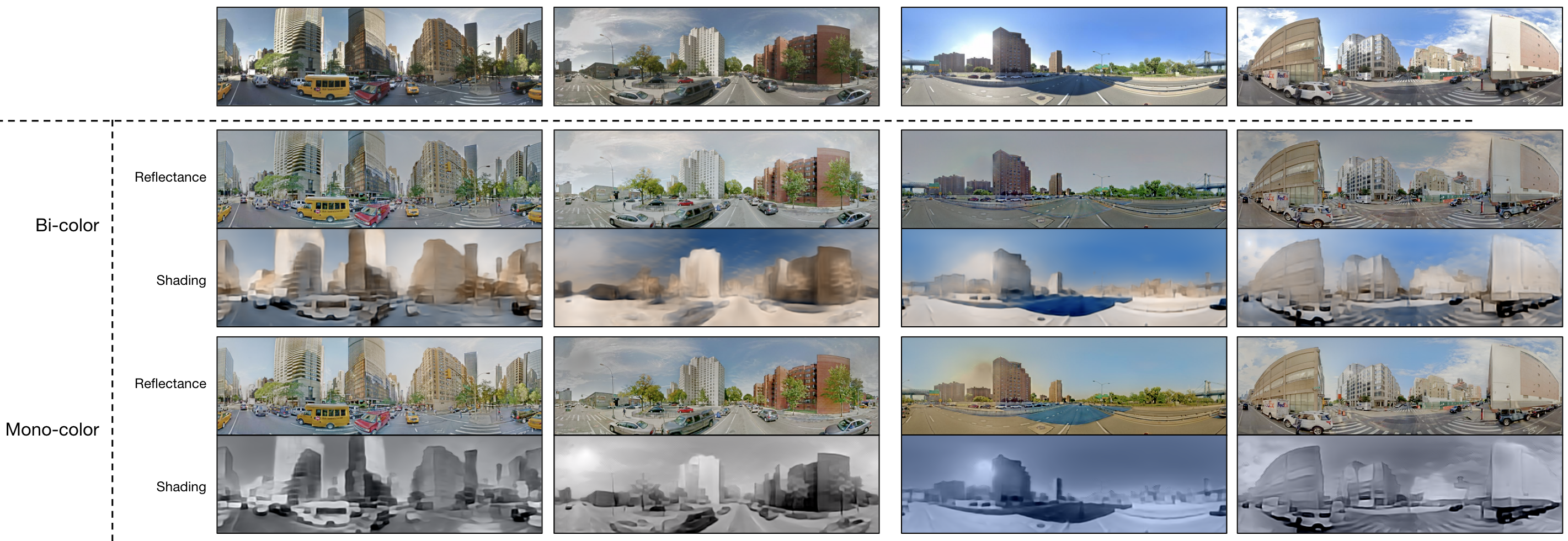}
    \end{center}
    \caption{
    We compare the benefits of the mono-color and bi-color shading assumption. We can see that the mono-color assumption fails to remove shadows completely. There are difficulties with white-balancing as well.}
    \label{fig:color-compare}
\end{figure*}

\begin{figure*}[t]
    \begin{center}
    \includegraphics[width=0.95\linewidth]{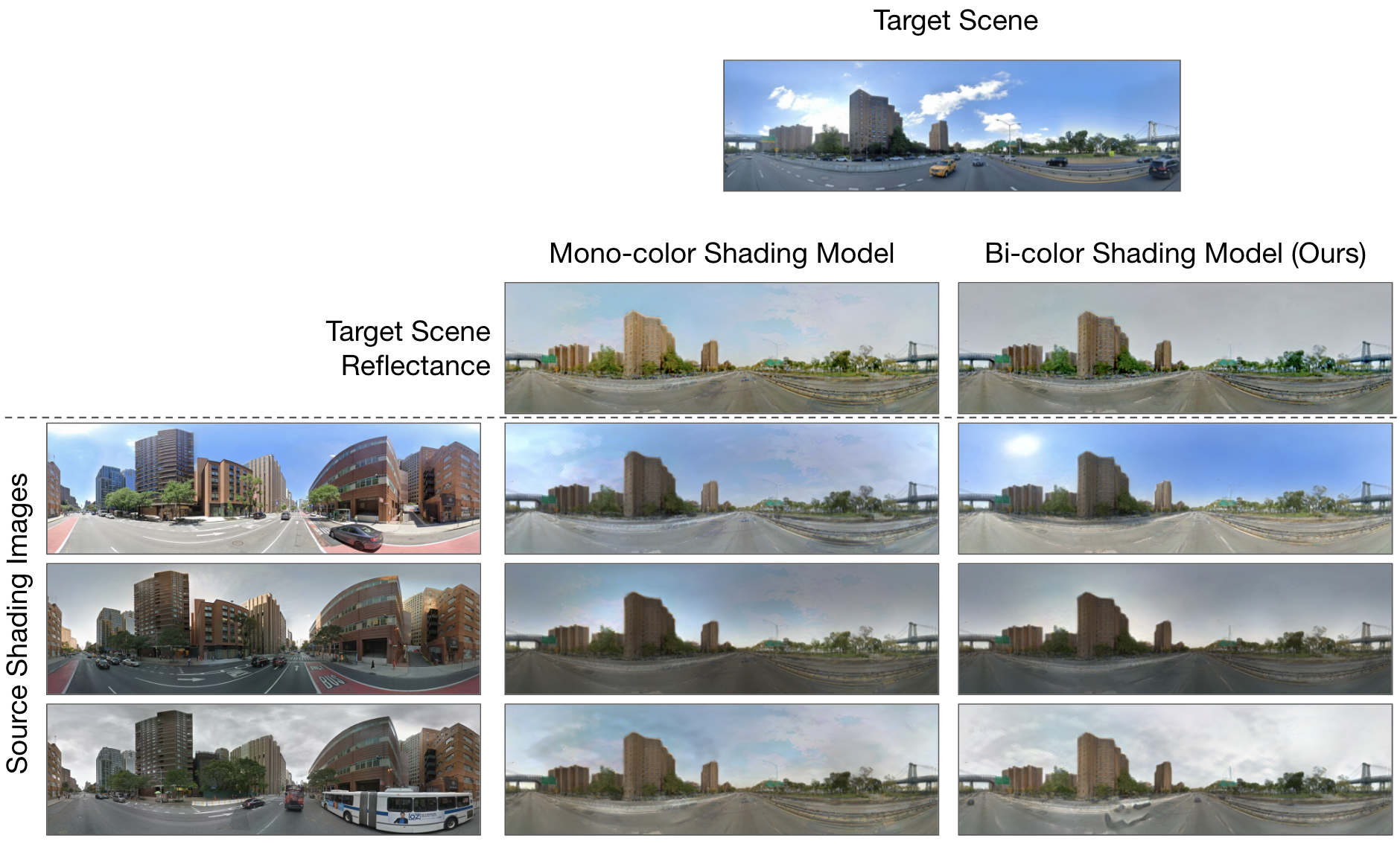}
    \end{center}
    \caption{
    For synthesizing new scenes by swapping lighting context, the bi-color shading model is a necessary improvement. The left column indicates the source weather we wish to copy from. The middle column shows a reconstruction under the mono-color shading. The right column is our bi-color shading. The mono-color fails to correctly synthesize any of the scenes correctly, partly because the reflectance captured is one of a blue sky instead of gray.}
    \label{fig:bicolor-swap-compare}
    \end{figure*}
\subsection{Bi-color shading}
\label{sec:bi-color-supp}
Our bi-color shading is very similar to Sunkavalli \etal~\cite{ftlv} proposed factorized timelapse model. In Sunkavalli \textit{etal}'s work, they decompose a spatio-temporal timelapse into a sky component, the scene illuminated by diffused sky-radiation, a binary shadow volume, and the scene illuminated by the sun. They alternate minimizing each component to arrive at the final decomposition for a given timelapse. While their results show a very good decomposition, their method suffers from the same issues with intrinsic images in that they cannot transfer appearances across space.

We show in Fig.~\ref{fig:generator} the different intermediate stages that are used to produce our bi-color shading output. Our SPADE generator $G$ takes in the factors and outputs the \textit{shading intensity} $\in \mathbb{R}^{H, W, 1}$ which is equivalent to the classic intrinsic image assumption of gray-scale shading. In addition, $G$ also estimates two global colors ($c_1$, $c_2$) $\in \mathbb{R}^3$ and a per-pixel mixing weight $M \in \mathbb{R}^{H, W}$. $c_1$ and $c_2$ are learned estimates of sunlight and skylight parameters; typically our model predicts various shades of blue and yellow-orange. These two are mixed using $M$ which is most similar to Sunkavalli \textit{etal}'s shadow volume. In particular each pixel in $M$ estimates the ratio of sunlight and skylight visible from the scene-point in the world. Our mixed colored representation from combining $c_1$, $c_2$, and $M$ defines the \textit{shading color} of the world. Our final outdoor illumination shading is like so:

\begin{equation}
    \log(S) + c_1 * M + c_2 * (1 - M)
    \label{eq:bi-color-output}
\end{equation}

In Fig.~\ref{fig:color-compare} we show qualitative difference between a mono-color ablation and the full model (bi-color). Note that cast shadow, especially in the third column, are removed from the bi-color reflectance. On the other hand, because the mono-color is incapable of modeling shadow volumes, they leave a strong blue tint behind. The bi-color shading also removes the diffuse sky radiation from the reflectance, leaving a clean plate background that is suitable for relighting as shown in Fig.~\ref{fig:bicolor-swap-compare}. The residual blue in the mono-color shading model degrades the swap reconstruction when transferring illumination from a different scene.

\subsection{Basis Spline Alignment}
We propose a novel procedure for dealing with alignment based on image congealing~\cite{congealing}. We find that alignment is an important problem to solve for producing high quality reflectances. We use these reflectances directly when synthesizing re-lighting scenes, as such alignment directly impacts the quality of our synthesized images. 

We'd like to re-iterate that while 3D reconstruction approaches like Martin-Brualla~\etal~\cite{TimelapseMiningSIGGRAPH15} and Meshry~\etal~\cite{Meshry_2019_CVPR} also have to solve for misalignment in input images, their approaches require many hours and hundreds of images to compute camera poses and dense 3D-reconstruction. Further these methods have no way to factorize completely unseen scenes. Our proposal enables all of this. We can deal with alignment from a few images and our encoder-decoder allows us to estimate intrinsic factors of unseen scenes from a single image.

Our approach initializes a set of free variables corresponding to control points $\Theta \in \mathbb{R}^{[8, 32, 2]}$ per image. The control points specify an $8, 32$ grid of horizontal and vertical deformations over an image. These deformations define a forward flow of pixels at the location of the $8, 32$ grid points in the image. To get a full and smoothly changing flow-field, we use a basis-spline (B-spline) to interpolate pixel deformations densely. We can apply the flow-field to warp an image. By warping all images in a stack using their respective $\Theta$, we get an aligned stack.

We use a cubic B-spline surface interpolation to compute the dense flow-field. The basis-spline surface interpolation is a generalization of the 1D B-spline to 2 dimensions corresponding to the height and width of control points. This is not related to the fact that we are learning horizontal and vertical deformations. Each horizontal and vertical deformation are independently computed B-spline surface interpolations.

Because the B-spline is differentiable, we can pass gradients through the B-spline and into the control points. While there exist other family of splines and differentiable warping (thin-plate splines for example), we found that the memory footprint and locally-constrained behavior of control points made B-splines the best candidate for interpolation.

We initialize the control point with noise, however the noise is small such that the initialization is essentially zero. This amounts to initializing the basis spline interpolation with the identity deformation.

\subsection{Loss Details}
Our primary loss operates on stacks of reflectance and shading outputs produced by our factorization model. Given a timelapse stack $\mathbf{I}$, we factorize the frames to their reflectance stack $\mathbf{R}$ and shading stack $\mathbf{S}$.

\medskip
\noindent
\textbf{Reflectance Consistency Loss.} Our reflectance consistency loss enforces the scene albedo to be consistent across time. We do this by minimizing the $L_1$ inconsistency between every pair of reflectance frames. This loss is used to jointly update alignment parameters and factorization.
\begin{equation}
    \mathcal{L}_\mathsf{RC} = \sum_i^8 \sum_{j=i+1}^8 ||\mathbf{R}_i - \mathbf{R}_j||_1
\end{equation}

\medskip
\noindent
\textbf{White light penalty.} In intrinsic images, there exists a fundamental ambiguity between log-reflectance and log-shading:

\begin{equation}
    \log(\mathbf{I}) = (\log(\mathbf{R}) - k) + (\log(\mathbf{S}) + k)
\end{equation}

where $k$ is an arbitrary channel-shift ambiguity of log-shading that affects the visualization of the components but \textit{does not} affect the resynthesis of the image. Typically this arbitrary shift plays a minor part in the decomposition as the expressiveness of $\log(S)$ is limited to be gray-scale intensity so applying normalization to visualize the components ignores the shift of $k$ in log-reflectance and log-shading. Additionally $\mathcal{L}_\mathsf{RC}$, and most approaches in intrinsic images, use shift invariant losses so $k$ cannot influence the optimization.

While we would normally not worry about this, our generator's bi-color shading is expressive enough to produce colored-components $c_1, c_2$. This means that $k$ ambiguity is an unconstrained color shift. For instance, if we pick an arbitrary color shift (in this case red) to be $k=\epsilon_\mathsf{red}$, the generator could learn to predict global color illuminants that have been red-shifted by $\epsilon_\mathsf{red}$. The result would be a log-shading output that appears red and log-reflectance output that appears cyan. When recombined though the ambiguity cancels and we get a regular image and an identical loss because $\mathcal L_\mathsf{RC}$ is invariant to $k$.

For visualization purposes, we impose a white-light loss $\mathcal{L}_\mathsf{WL}$ that enforces the average colored illumination $c_1, c_2$ across time to be white. This effectively encourages the shading generator $G$ to prefer solutions where the color shift ambiguity $k$ is white. As a result we get log-reflectance that appear illuminated under white-light. We take the bi-color augmentation described in Sec.~\ref{sec:bi-color-supp} $c_1 * M + c_2 * (1 - M)$, denoted as shading color in Fig.~\ref{fig:generator}, to be $\mathbf{B} \in \mathbb{R}^{[8, H, W, 3]}$ and $\mathbf{B}_i$ to be the bi-color component for frame $i$.

\begin{equation}
  \mathcal{L}_\mathsf{WL} = \sum ||\sum_i^8 \mathbf{B}_i||_1
\end{equation}
\medskip
\noindent
\textbf{Misc loss.} 
We adopt a standard GAN setup. We use a patch discrminator with 1 scale and 4 layers~\cite{pix2pix2016,wang2018pix2pixHD}. Our patch discriminator's only input is the stack reconstruction pixels that have been synthesized from the average reflectance and the predicted shading. Please refer to Fig.~\ref{fig:decoder-disc} for more detail.

We use a hinge adversarial loss $\mathcal{L}_\mathsf{GAN}$. $\mathcal{L}_\mathsf{GAN}$ appropriately switches between the following generator and discriminator loss when computing gradients of their respective networks. $X$ represent the real images, $D(X)$ represents the discriminator logits, and $F(X)$ represents the encoder-decoder stack reconstruction:

\begin{equation}
    \mathcal{L}_\mathsf{Disc} = \max(1 + D(X), 0) + \max(1 - D(F(X)), 0)
\end{equation}

\begin{equation}
    \mathcal{L}_\mathsf{Gen} = -D(F(X))
\end{equation}

We also use a feature-matching loss $\mathcal{L}_\mathsf{FM}$ which guides the encoder-decoder to produce images $F(X)$ that are similar to $X$. This is done by matching the intermediate activations of the discriminator between $X$ and $F(X)$. Let $D_i(*)$ refer to the activations of the i-th layer.

\begin{equation}
    \mathcal{L}_\mathsf{FM} = \sum_i ||D_i(X) - D_i(F(X))||_1
\end{equation}

Lastly we include a perceptual loss~\cite{perceptualloss} $\mathcal{L}_\mathsf{VGG}$ which also guides the encoder-decoder to produce images $F(X)$ that are ``perceptually" similar to the real image $X$. This is enforced using $L_1$ loss between real and generated samples' VGG-19 features. We re-use the implementation from~\cite{Chen_2017_ICCV}.

\subsection{Training Details}
We used V100 GPUs with asynchronous gradient updates over a total of $100,000$ stacks each consisting of 8 panoramic images.

Our training stacks are augmented by horizontally translating all their constituent panoramas by the same amount after alignment warping, but before decomposing. This is equivalent to randomly rotating the canonical heading and prevents the model from overfitting to the natural pattern of the sun's position (e.g at higher latitudes in the northern hemisphere, the sun is typically not observed in the geographically north part of the sky). 

For both our factorization and discriminator update, we used the default Adam~\cite{kingmaadam} with learning rate 0.0001 and $\beta_1=0$ which was found to work well in SPADE~\cite{taesung}. For learning our warp parameters $\Theta$, we used a lazy implementation of Adam \footnote{\texttt{tf.contrib.opt.LazyAdamOptimizer}} that's optimized for applying efficient sparse updates.

\section{Additional Results}
\subsection{Sun Azimuth Evaluation}
\setlength{\tabcolsep}{4pt}
\begin{SCtable}[\sidecaptionrelwidth][t]
\begin{minipage}[t!]{0.50\linewidth}
\small
\centering 
\resizebox{\columnwidth}{!}{
\begin{tabular}{lcc}
\toprule
Model & Test-GSV  & Laval\\
\midrule
Ours       & 0.806 (9.2\textdegree) & 0.771 (9.6\textdegree) \\ %
Supervised Azimuth Encoder & \textbf{ 0.864 (7.92\textdegree)} & 0.831 (9.3\textdegree) \\
Deep Outdoor Illumination~\cite{deepskymodel} & --- & N.A \textbf{(4.59\textdegree)}\\
\hline
\end{tabular}
}
\end{minipage}\hfill
\begin{minipage}[t!]{0.5\linewidth}
\caption{Estimating sun azimuth from panoramas. We report the average cosine similarity between prediction and ground truth (higher is better). In parenthesis, the median angular error (lower is better). 
}
\label{table:azimuth}
\end{minipage}
\end{SCtable}
\setlength{\tabcolsep}{4pt}

We evaluate the goodness of our unsupervised azimuth estimation module by comparing our unsupervised estimates with the true azimuth heading on two test datasets: GSV-TM and Laval Outdoor HDR~\cite{deepskymodel}. The true azimuth for GSV-TM is computed using solar angle equations from the GPS and date-time metadata. Laval panoramas are annotated with azimuth estimated from computing connected components of the brightest pixel.

In order to measure correctness in azimuth estimation, we compute the cosine distance between the predicted and true azimuth angles. Because our azimuth representation is an unsupervised embedding learned by a neural network, the relationship between the output of the encoder and the ground truth sun azimuth angle is ambiguous up to a constant rotation. Therefore, we estimate a rotation of our azimuth representation over a validation set that maximizes the cosine similarity between our finetuned rotated prediction and the real angle of the sun. 

We show the full results in Table~\ref{table:azimuth}, comparing against a fully supervised azimuth encoder as well as a supervised baseline method~\cite{deepskymodel} on the Laval dataset. We record the average cosine similarity and median angular error (shown in parenthesis) between the prediction and ground truth.

\subsection{Alignment Results:}
\begin{figure*}[t]
    \centering
    \includegraphics[width=0.96\linewidth]{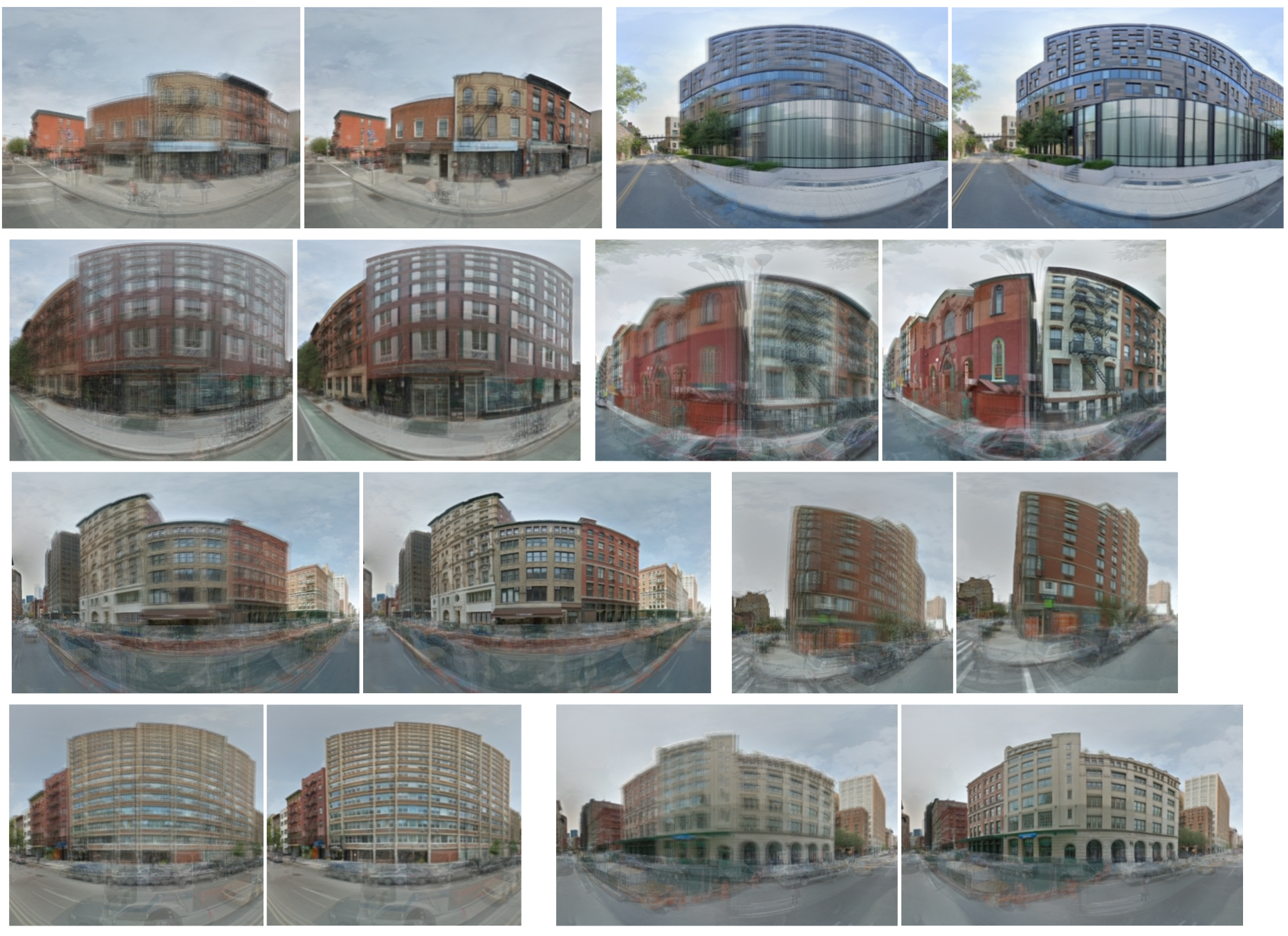} 
    \caption{Additional alignment result. For each pair we show the unaligned average on left and the output of our alignment algorithm on right. These alignments are computed at test-time when the factorization weights are frozen. Given a stack we take ten gradient descent optimization of alignment parameters $\Theta$.} 
    \label{fig:supp-align-results}
\end{figure*}

\begin{figure*}[t]
    \centering
    \includegraphics[width=0.96\linewidth]{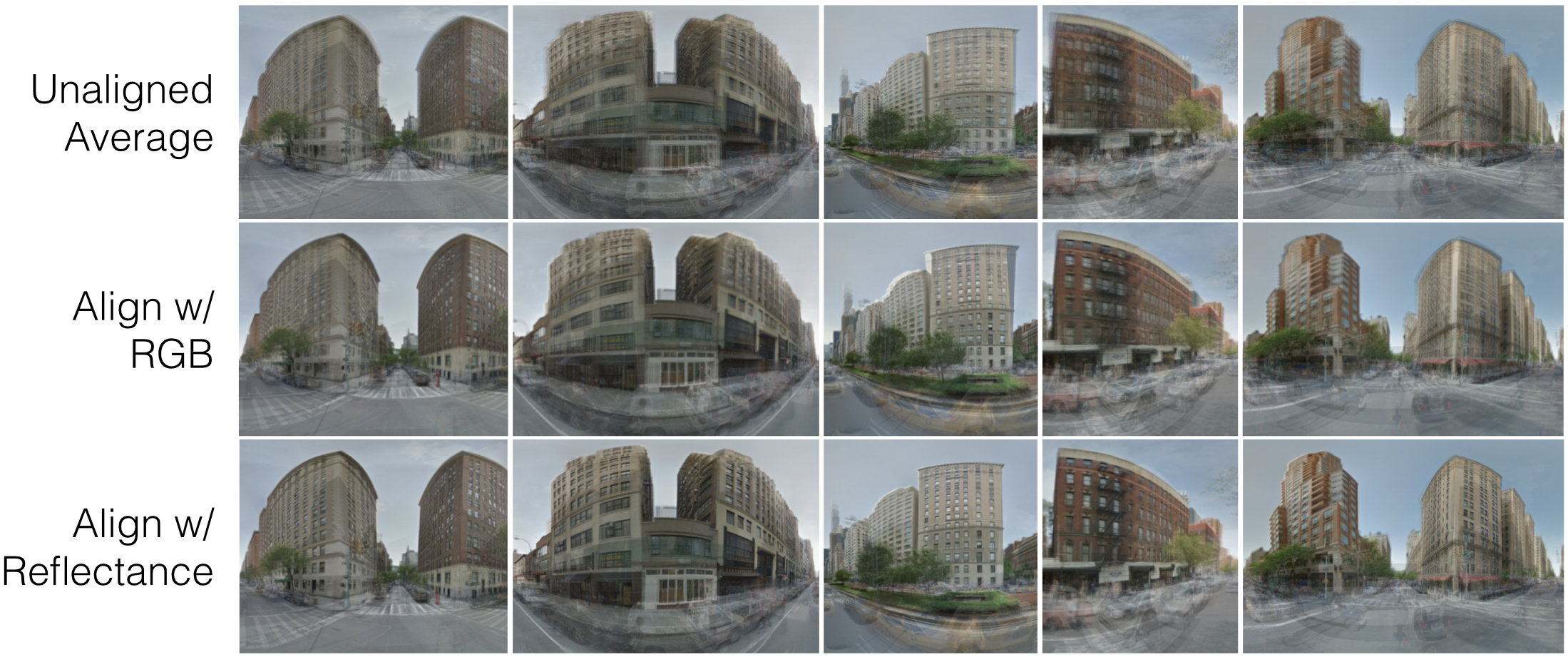} 
    \caption{We show two versions of the basis-spline alignment process on the unaligned stacks in the top row. The middle row shows our resulting alignment average after optimizing for aligning the original RGB pixels. This is equivalent to breaking the alignment-factorize loop by aligning the stacks as a pre-processing step. The bottom row shows our full-model with the alignment-factorize loop. These examples show that aligning with reflectance produces sharper building textures than aligning with RGB. These examples were selected to highlight the discrepancy in alignment results, but even for the median case there are many subtle misalignment problems that aren't immediately obvious.} 
    \label{fig:supp-align-compare}
\end{figure*}

In Fig.~\ref{fig:supp-align-results} we show additional alignment on test stacks. The misaligned stacks highlight the importance of solving alignment for producing high quality images. While we proposed solving for alignment with the factorized reflectance, one could choose to break the feedback loop and solve for alignment using the original RGB pixels. This is equivalent to first solving for alignment as preprocessing to the encoder-decoder.

We show in Fig.~\ref{fig:supp-align-compare} that aligning on the original RGB image stack results in poorer alignment than our proposed process. While we did not validate this, we suspect that attempting RGB alignment on even smaller sized stack would result in even more poor results.

\subsection{Beyond NYC}
\begin{figure}[t]
    \begin{center}
    \includegraphics[width=0.9\linewidth]{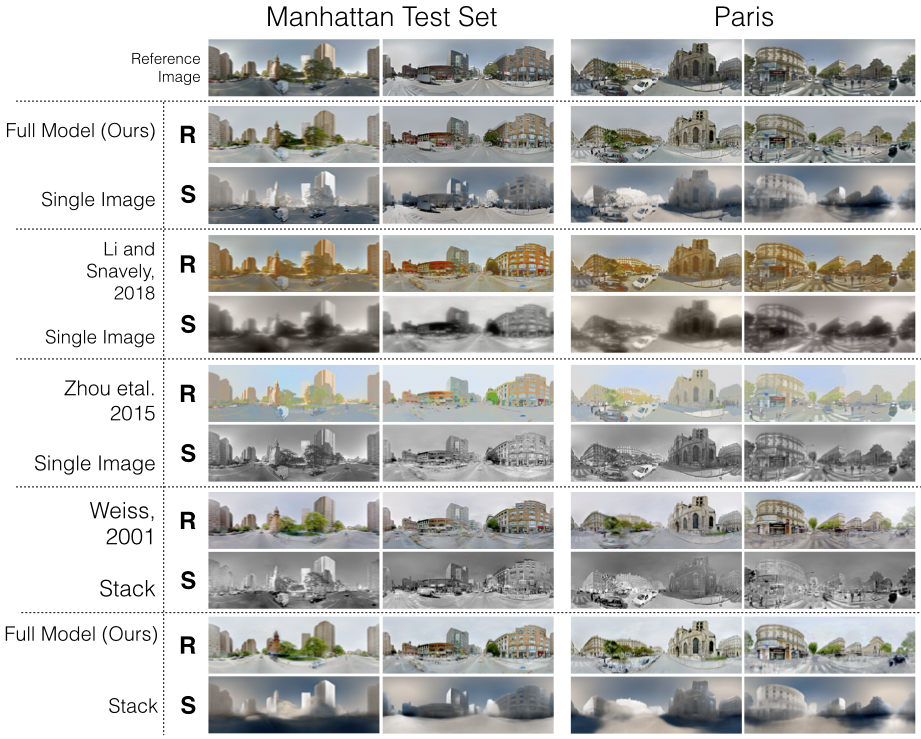}
    \end{center}
    \caption{Additional results from main Fig.\ 6. In addition we also show a stack decomposition comparison between our approach and Weiss's MLE Intrinsics~\cite{weiss2001intrinsics}.} 
    \label{fig:nyc_paris}
\end{figure}

\begin{figure}[t]
    \begin{center}
    \includegraphics[width=0.9\linewidth]{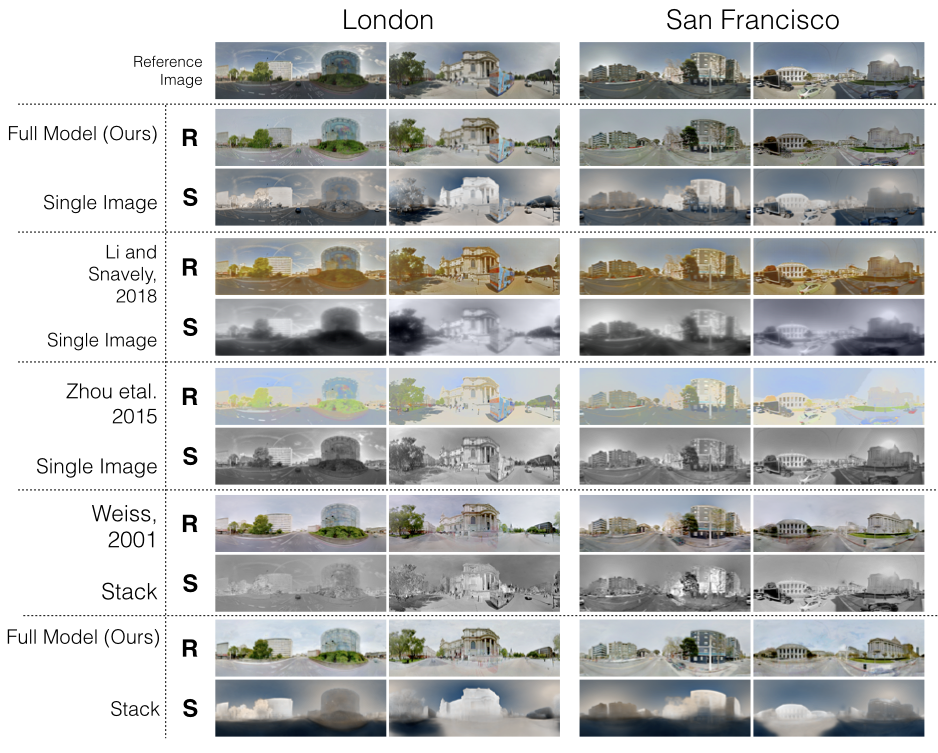}
    \end{center}
    \caption{More decomposition on city scenes from beyond NYC. We show that our factorization generalizes beyond location as shown by our results on London and San Francisco.}
    \label{fig:london_sf}
\end{figure}

Majority of our results were shown on imagery from the test set of NYC GSV-TM. We briefly showed earlier that our factorization works for scenes beyond NYC with intrinsic image decompositions of Paris, London, and Laval Outdoor HDR.

We include additional decomposition results for each city and also San Francisco in Fig.~\ref{fig:london_sf}. In addition to the decomposition from the main submission, we also show a stack decomposition comparison with Weiss's MLE Intrinsics~\cite{weiss2001intrinsics} and our model on stacks. In the stack decomposition, our full-model has a better signal for removing moving objects like cars and people from both the reflectance and geometry. 

\section{Applications}
\subsection{Transferring only Lighting Context}
\begin{figure}[t]
    \begin{center}
    \includegraphics[width=1.0\linewidth]{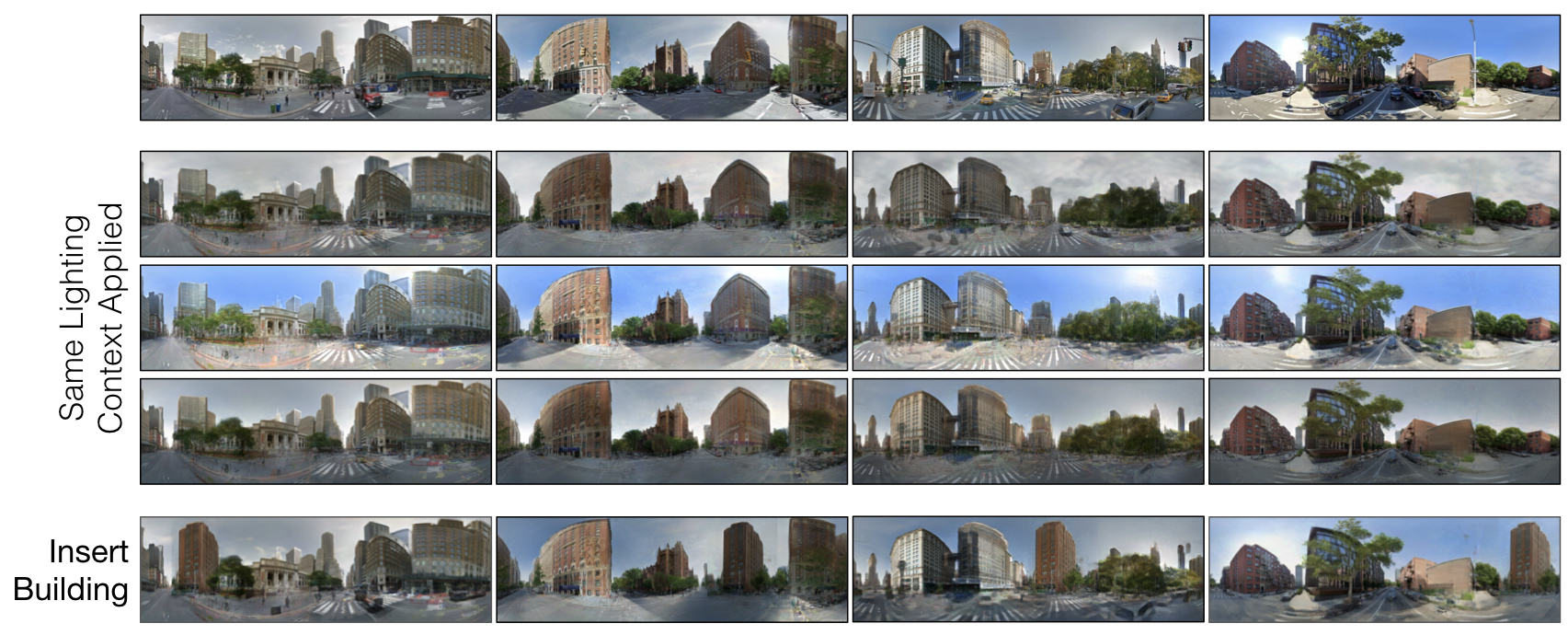}
    \end{center}
    \caption{We show results for applications enabled by our factorization. In the top row we show various scenes we wish to manipulate. The middle section show these scenes relit with a consistent lighting context but original azimuth location. This indicates that we have disentangled sun azimuth from lighting context. In the bottom row we show transplanting a building into the world and updating the lighting realistically.}
    \label{fig:supp-results}
\end{figure}

So far we have visualized transferring the whole \textit{illumination descriptor} and manipulating just the azimuth representation $\varphi$. To fully demonstrate disentanglement, we also show manipulating just the \textit{lighting context} in Fig.~\ref{fig:supp-results}(middle). Here we show different scenes with different sun azimuth. For a row, we transfer the same lighting context $L$ while preserving the original sun azimuth. Note how the same bright blue sky is synthesized, but the sun location matches the original scene's sun azimuth.

\subsection{Editing Scene Geometry}
\begin{figure}[t]
\begin{center}
\includegraphics[width=0.7\linewidth]{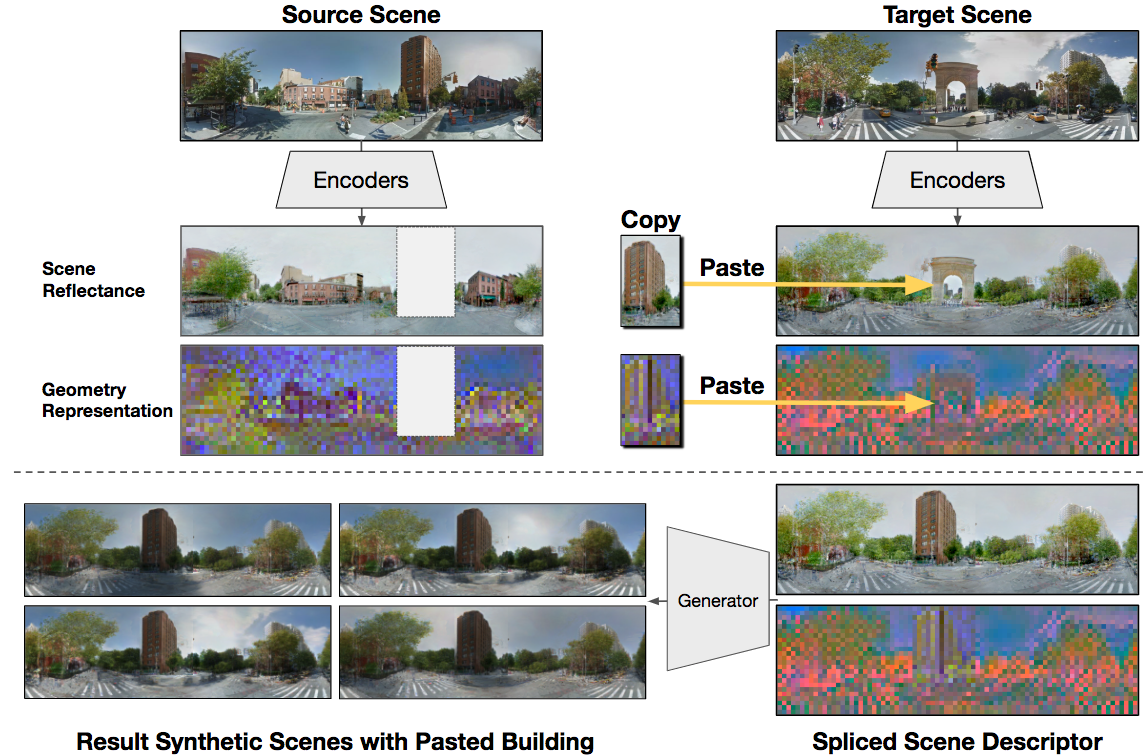}
\end{center}
   \caption{Our factorized representation lets us copy parts of the scene descriptor and paste them into new scenes before generating realistic-looking modified panoramas. Our synthetically inserted brown building is seamlessly integrated into the target scene. 
   }
\label{fig:splice}
\end{figure}

We show the procedure of inserting objects into different scenes in Fig.~\ref{fig:splice}. This underlying process is how we can synthetically transplant buildings into new scenes. Additionally the newly synthesized geometry code forms a new scene for which all previously defined factored transformations can be applied. We show scene rotations of the newly spliced building on the attached webpage.

\subsection{Hyperlapse Synthesis}
\begin{figure*}[t]
    \begin{center}
    \includegraphics[width=0.95\linewidth]{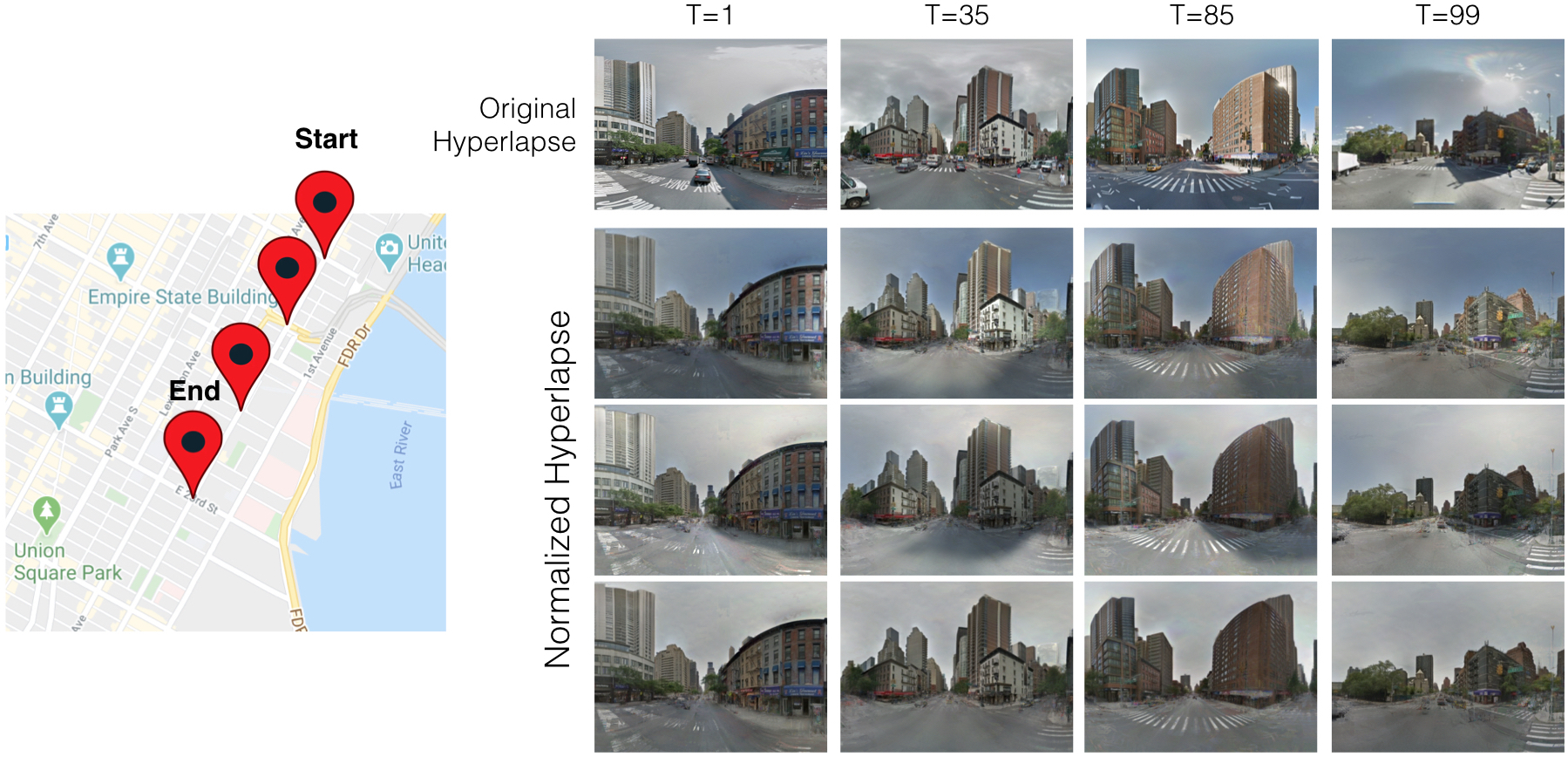}
    \end{center}
    \caption{We show a hyperlapse drive down Second Avenue, Manhattan. In the original hyperlapse, the illumination changes frequently resulting in a jarring experience. We show the process of fixing the illumination to make the hyperlapse weather consistent. The full video can be found attached in the supplemental folder. 
    }
    \label{fig:hyperlapse}
\end{figure*}

We show a hyperlapse synthesis through a section of New York City. In particular the original images often have uncontrollable changes in lighting due to images coming from distinct capture times, resulting in a jarring experience. Using our factorization method, we can normalize for lighting and drive smoothly through Manhattan. Video of the original hyperlapse and adjusted one can be found on the attached webpage.

\section{Failure Cases}
\begin{figure}[t]
    \begin{center}
    \includegraphics[width=0.8\linewidth]{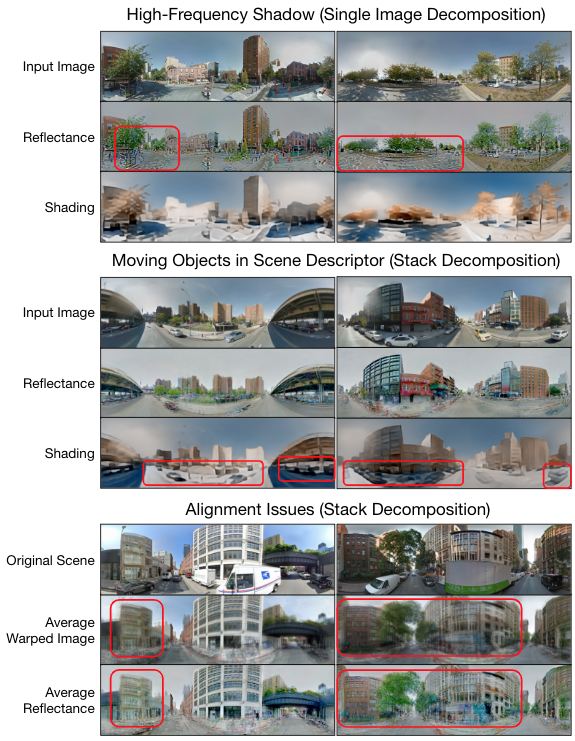}
    \end{center}
    \caption{We show three common failure of our model as well as whether it is associated with a single-image or stack decomposition. In the first row we show how our decomposition fails to correctly remove high-frequency shadows from reflectance like ones left by branches. In the second row we show failures to remove transient objects like cars from the scene descriptor. This results in the synthesis of ``ghost" cars. Finally we show a failure of our alignment module. The average images are poorly align, resulting in a poor estimate of the average reflectance.}
    \label{fig:failure}
\end{figure}

We show some common failure modes of our factorization in Fig.~\ref{fig:failure}. For single image decompositions, the single biggest source of failures results from poor estimates of our scene descriptors (reflectance and geometry). In the first row, the single image decomposition struggles to correctly synthesize high-frequency shadows cast by branches. This suggests two things: (1) that multiple views improve the decomposition results by letting the network average out poor geometry and reflectance estimates and (2) the nature of our compressed factorization forces intricate shading interactions like branch shadows to be encoded in the scene descriptor.

The next common failure mode is a result of ghosting of transient objects like cars. While under our current factorization there's no intuitive place to encode cars because they represent changes in the underlying scene geometry that are not permanent. The network learns to average out moving objects in reflectance and attempts to best reconstruct them in the shading images, resulting in wispy gray-scale cars. The second row shows examples of these ghost cars.

Another common failure mode is a poor alignment of timelapse images. Even though images are within a 0.4m radius circle, there are certain scenes where a the basis spline cannot correctly align images due to exceptionally bad parallax. In the last row of Fig.~\ref{fig:failure} we show timelapses with poor alignment parameters, resulting in an equally bad estimate of reflectance. 

Based on results from image congealing~\cite{congealing}, congealing over larger sets of images results in better chance of aligning images by smoothing the optimization surface. Therefore two possible solutions exists to make alignment better: (1) we can use a smaller baseline to decrease the adverse impact of parallax and (2) we can use more images per stack to smooth the optimization surface when aligning images.
\end{document}